\PassOptionsToPackage{dvipsnames,table}{xcolor}

\documentclass[10pt,twocolumn,letterpaper]{article}

\usepackage[pagenumbers]{iccv} 

\usepackage{graphicx}
\usepackage{caption}
\usepackage{subcaption}

\usepackage{tabu}
\usepackage{tikz}

\newcommand\blfootnote[1]{%
  \begingroup
  \renewcommand\thefootnote{}\footnote{#1}%
  \addtocounter{footnote}{-1}%
  \endgroup
}

\usepackage{bm}

\usepackage[table]{xcolor}
\usepackage{bbding}
\usepackage{pifont}
\usepackage{wasysym}
\usepackage{amssymb}
\definecolor{cvprblue}{rgb}
{0.21,0.49,0.74}
\definecolor{iccvblue}{rgb}{0.21,0.49,0.74}
\definecolor{deepskyblue}{rgb}{0.0, 0.75, 1.0}
\usepackage[pagebackref,breaklinks,colorlinks,allcolors=deepskyblue]{hyperref}

\newcommand*\colourcheck[1]{%
  \expandafter\newcommand\csname #1check\endcsname{\textcolor{#1}{\ding{52}}}%
}
\colourcheck{blue}
\colourcheck{green}
\colourcheck{red}

\usepackage{multirow}

\usepackage{arydshln}

\usepackage{algorithm}
\usepackage[noend]{algpseudocode}

\definecolor{FFABA8}{HTML}{e3ebfc}
\colorlet{Light}{FFABA8}
\newcommand{\CC}[1]{\cellcolor{Light}}

\definecolor{mygray}{gray}{0.6}

\newcommand{\mbf}[1]{\mathbf{#1}}

\newcommand{\s}{\mathbf{s}}
\newcommand{\y}{\mathbf{y}}

\definecolor{demphcolor}{RGB}{144,144,144}
\newcommand{\demph}[1]{\textcolor{demphcolor}{#1}}

\definecolor{color1}{HTML}{7735bd}
\definecolor{color2}{HTML}{398cb3}
\definecolor{color3}{HTML}{18a370}

\definecolor{nicegreen}{rgb}{0.1, 0.6, 0.2}

\def\paperID{13191} 
\def\confName{ICCV}
\def\confYear{2025}

\newcommand{\customsubsection}[1]{%
  \par
  \pagebreak[2]%
  \refstepcounter{subsection}%
    \everypar={%
      {\setbox0=\lastbox}
      \addcontentsline{toc}{subsection}{%
        {\protect\makebox[0.3in][r]{\thesubsubsection.} \hspace*{3pt}#1}}%
      \textbf{\thesubsection\space\space{#1}\space\newline}%
      \everypar={}%
    }%
  \ignorespaces
}

\newcommand{\customsubsubsection}[1]{%
  \par
  \pagebreak[2]%
  \refstepcounter{subsubsection}%
    \everypar={%
      {\setbox0=\lastbox}
      \addcontentsline{toc}{subsubsection}{%
        {\protect\makebox[0.3in][r]{\thesubsubsection.} \hspace*{3pt}#1}}%
      \textbf{\thesubsubsection\space\space{#1}\space}%
      \everypar={}%
    }%
  \ignorespaces
}

\def\ourapproach{\textsc{EgoAdapt}\xspace}
\def\ourpolicy{TeMPLe\xspace}

\title{\textsc{EgoAdapt}: Adaptive Multisensory Distillation and Policy Learning \\ for Efficient Egocentric Perception}

\author{
Sanjoy Chowdhury$^{1,2,\dagger}$ \quad Subrata Biswas$^{2,3\dagger}$ \quad Sayan Nag$^4$ \quad Tushar Nagarajan$^5$ \\
Calvin Murdock$^2$ \quad Ishwarya Ananthabhotla$^2$ \quad Yijun Qian$^2$ \\
Vamsi Krishna Ithapu$^2$ \quad Dinesh Manocha$^1$ \quad Ruohan Gao$^1$ 
\vspace{1mm} \\
$^1$University of Maryland, College Park \quad
$^2$Meta Reality Labs \quad
$^3$Worcester Polytechnic Institute \\
$^4$University of Toronto \quad
$^5$FAIR, Meta AI
\vspace{0.5mm} \\
\texttt{\footnotesize\{sanjoyc, dmanocha, rhgao\}@umd.edu} \quad
\texttt{\footnotesize \{tusharn, cmurdock, ishwarya, ithapu\}@meta.com} \\
\texttt{\footnotesize sayan.nag@mail.utoronto.ca} \quad \texttt{\footnotesize sbiswas@wpi.edu} \\
}
\begin{document}
\maketitle

\begin{abstract}
\blfootnote{$ ^\dagger$ Work done during internship at Meta Reality Labs. }
Modern perception models, particularly those designed for multisensory egocentric tasks, have achieved remarkable performance but often come with substantial computational costs.
These high demands pose challenges for real-world deployment, especially in resource-constrained environments. 
In this paper, we introduce \ourapproach, a framework that adaptively performs cross-modal distillation and policy learning to enable efficient inference across different egocentric perception tasks, including egocentric action recognition, active speaker localization, and behavior anticipation. Our proposed policy module is adaptable to task-specific action spaces, making it broadly applicable. 
Experimental results on three challenging egocentric datasets---EPIC-Kitchens, EasyCom, and Aria Everyday Activities---demonstrate that our method significantly enhances efficiency, reducing GMACs by up to $89.09\%$, parameters up to $82.02\%$, and energy up to $9.6 \times$, while still on-par and in many cases outperforming, the performance of corresponding state-of-the-art models.

\end{abstract}

\begin{figure}[t]
\centering\includegraphics[width=\columnwidth]{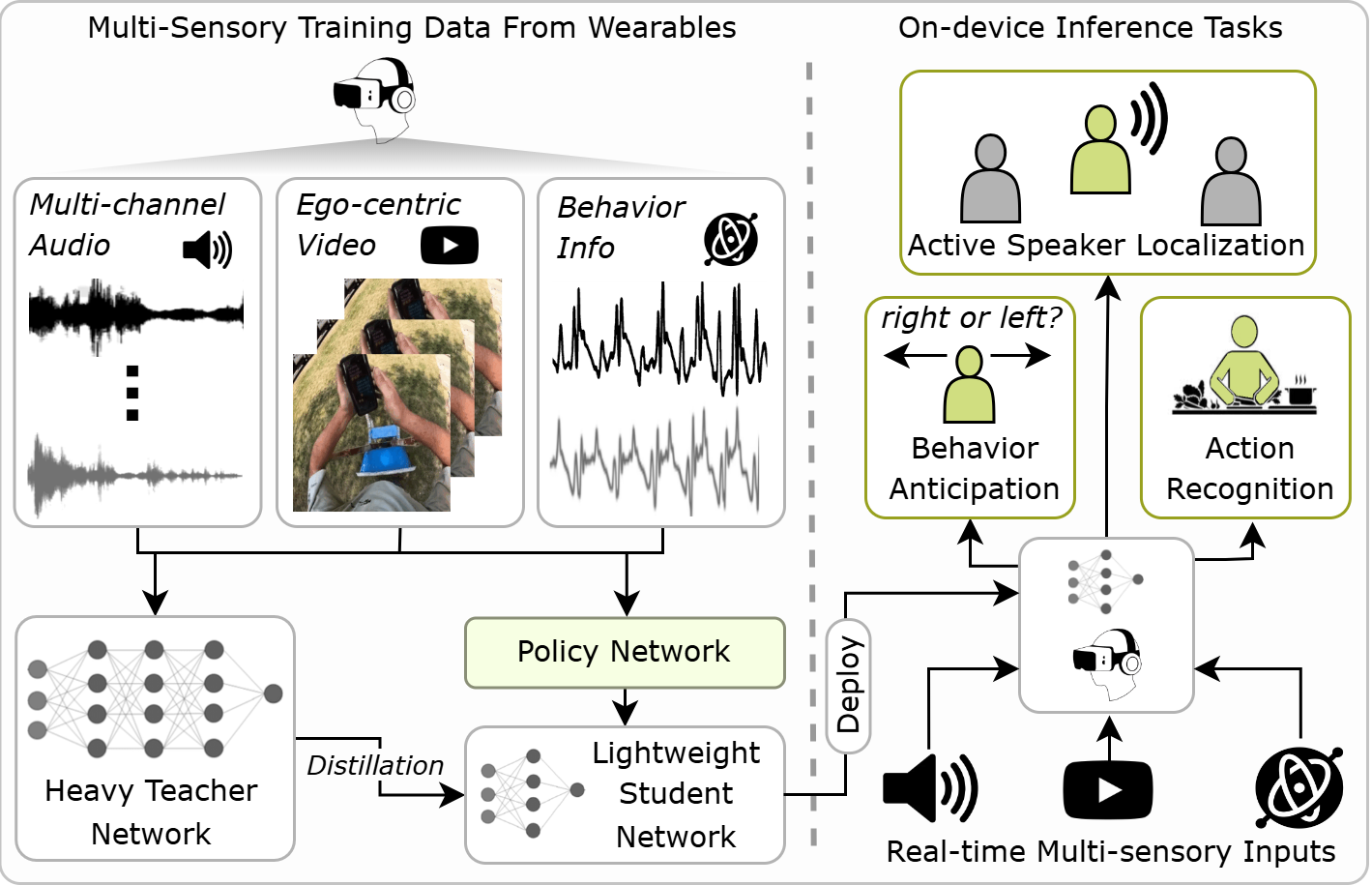}
  \vspace{-0.2in}
  \caption{\textbf{We introduce \ourapproach, a unified framework combining cross-modal distillation and policy learning} for efficient multisensory egocentric perception. It optimizes the use of constituent modalities for efficient inference and is adaptable to various egocentric perception tasks, including action recognition, active speaker localization, and behavior anticipation.}
  

  
  \label{teaser}
  \vspace{-1em}
\end{figure}

\vspace{-5mm}

\section{Introduction}

Recent developments in augmented reality (AR) and virtual reality (VR) are revolutionizing the way people interact with the digital world. 
These systems often rely on the ability to recognize user behaviors---such as eye gaze, head motion, and hand-object interactions---and understand the surrounding contexts using multisensory data streams.
However, achieving real-time inference in such applications remains a challenge due to the substantial computational demands of state-of-the-art egocentric perception models. Efficient computation is essential not only for managing resource constraints, but also to enable faster response times, reduced latency, and improved overall system performance, all of which are crucial for delivering an engaging and interactive AR/VR experience. 

Despite significant progress in egocentric video understanding~\cite{datta2022episodic, damen2022rescaling, grauman2022ego4d, fan2021multiscale, he2019rethinking, omnivore, li2022egocentric, liu2022joint, patrick2021keeping, plizzari2022e2, jia2024avgraph, perceptiontest, sigurdsson2018charades}, most current approaches still rely on computationally expensive clip-based models, similar to those used for conventional third-person view videos \cite{park2022per, yang2021beyond, pei2023clipping}. 
While effective in analyzing and understanding users and their environments, these models often fail to account for the necessity of resource efficiency, particularly in scenarios where power, memory, or thermal constraints are limiting factors.

A key observation is that many egocentric perception models process multiple sensory modalities---such as RGB video, multi-channel audio, and behavioral data---simultaneously, even when all are not necessary for accurate inference, as illustrated in Fig.~\ref{teaser}. 
For instance, in the egocentric active speaker localization task \cite{jiang2022egocentric}, which aims to detect the spatio-temporal locations of active speakers, different modalities can be selectively employed: when multiple people are visible, visual cues such as gestures, facial expressions, and body motions can indicate who is speaking, whereas when speakers are outside field of view, multi-channel audio cues can be used to determine their locations \cite{tourbabin2019direction}. Similarly, auxiliary data, such as pose information from IMU sensors, can provide useful context while being computationally more efficient than processing high-resolution video streams. This suggests that adaptive use of different modalities can potentially significantly reduce computational overhead while maintaining performance.



To achieve this, we argue that two key strategies are essential. First, lightweight models should be employed for processing each modality, replacing the powerful yet costly models typically used in the literature. Second, there is a need for an adaptable mechanism to selectively utilize  
necessary modalities based on context and varying computational constraints. Specifically, this involves devising a policy-driven approach that adaptively selects the appropriate modality---such as video frames and audio channels---at the right time. While prior work has explored either model distillation \cite{listentolook, tian2023view, yang2022efficient, radevski2023multimodal} or adaptive modality selection in isolation \cite{adamml, radevski2023multimodal}, a unified approach that integrates both strategies is crucial for maximizing efficiency.

Model distillation reduces computational cost but remains static, failing to adapt to varying task demands. Conversely, adaptive modality selection dynamically allocates resources but often relies on expensive models, limiting overall gains. By jointly learning model distillation and policy optimization,  the approach benefits from a synergistic integration where the efficiency of a distilled model is complemented by the adaptability of policy optimization. This joint training enables the system to continuously refine its resource allocation strategy while maintaining low computational overhead, thereby achieving a balance between performance and efficiency that neither technique can reach alone. For example, \ourapproach achieves an improvement of $\sim$ 22\% on ASL compared to when joint distillation and policy learning is not employed.


To summarize, our main contributions are: 
\begin{itemize}
\item We propose a unified framework that harnesses the strengths of both cross-modal distillation and policy learning to achieve the optimal balance between performance and efficiency. 

\item We efficiently train a multi-modal policy network jointly with the distillation model using standard backpropagation through Gumbel-Softmax sampling, adapting to various egocentric tasks 
by expanding its action space. 

\item We validate our approach on three challenging datasets and egocentric perception tasks, achieving up to $89.09\%$ reduction in GMACs, $82.02\%$ reduction in parameters, and up to $9.6 \times$ reduction in energy, while still on-par and in many cases outperforming, the performance of corresponding state-of-the-art models.

\end{itemize}

\section{Related Work}

\noindent{\textbf{Efficient Video Understanding.}} The increasing demand for on-device applications has driven interest in efficient video recognition. Consequently, recent works are focusing on creating lightweight architectures that achieve high video recognition performances while conserving computational resources. Several studies focus on designing lightweight architectures \cite{feichtenhofer2020x3d, herath2020ronin, kondratyuk2021movinets, tran2018closer, zolfaghari2018eco} by reducing 3D CNN operations across densely sampled frames. 

Another approach to efficiency focuses on adaptively selecting video content. Some methods reduce temporal redundancy by selectively processing video clips \cite{korbar2019scsampler}, frames \cite{frameexit, meng2020ar}, and feature channels \cite{adafuse}, while others address spatial redundancy by dynamically selecting smaller but important regions \cite{wang2021adaptive, wang2022adafocus}. Another work \cite{wang2022efficient} selects tokens in video transformers across spatial and temporal dimensions. Complementary to these methods, our method optimally leverages constituent modalities by learning an adaptable strategy that can efficiently switch between and guide each other to perform various downstream egocentric tasks.



\begin{figure*}[t]
\centering
  \includegraphics[width=.98\textwidth]{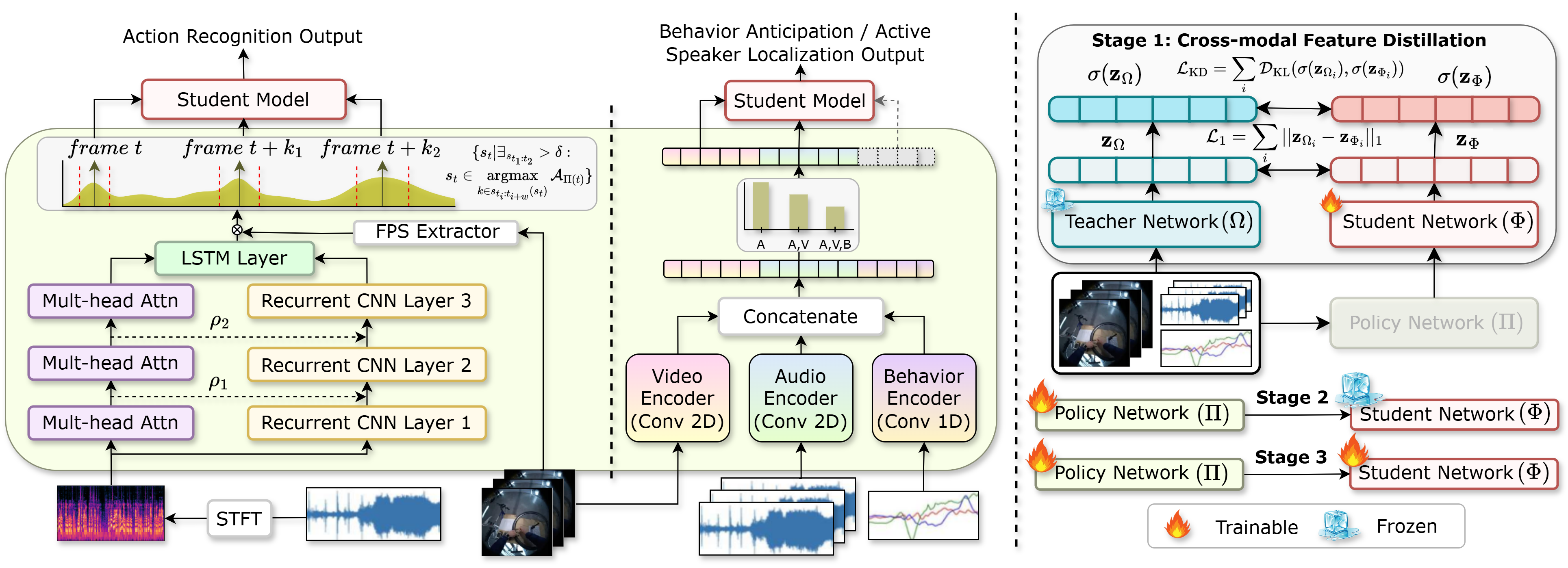}
  \vspace{-0.05in}
  \caption{\textbf{Illustration of \ourapproach.} Our framework consists of two main components, a lightweight policy module \emph{$\Pi$} and a distillation module \emph{$\Phi$} composed of different sub-networks that are trained jointly (via late fusion with learnable weights) for various egocentric tasks, including action recognition, active speaker localization, and behaviour anticipation. The policy module is adaptable, dynamically selecting the optimal modality, video frame, and audio channel based on the downstream task
  to balance performance and efficiency. The entire pipeline is trainable jointly over multiple stages (right side). In Stage 1, the policy module is disabled. FPS: Frames per second. }
  \label{main_diagram}
  \vspace{-1em}
\end{figure*}




\vspace{0.05in}

\noindent{\textbf{Multimodal Distillation.}} Knowledge distillation transfers knowledge from an expensive `teacher' model to a lightweight `student' model \cite{knowledgedistillation}. Approaches fall into three categories based on the knowledge type: responses, features, and relations. Response-based methods align the student’s predictions with the teacher's \cite{beyer2022knowledge, zhao2022decoupled}, feature-based methods match intermediate features \cite{shu2021channel, wang2019distilling}, and relation-based methods align inter-sample relationships between the teacher and student networks \cite{tian2019contrastive, park2019relational}.


Recent work explores multimodal distillation in various ways: transferring from a RGB model to a flow or depth model \cite{garcia2018modality, gupta2016cross}, from a 3D model to a 2D model \cite{liu20213d}, or from a visual model to audio model \cite{gan2019self, aytar2016soundnet}. Listen To Look \cite{listentolook} incorporates both clip subsampling and video-to-audio distillation for fast activity recognition in third-person videos. EgoDistill \cite{egodistill} further explores the relationship between the camera wearer’s head motion and RGB signals to guide the distillation process. Different from these methods, we present a novel way to effectively combine policy and distillation for tackling different egocentric tasks. 


\vspace{0.05in}

\noindent{\textbf{Adaptive Computation.}} 
Many adaptive computation methods are proposed to improve efficiency \cite{bengio2015conditional, bengio2013estimating, veit2018convolutional, wang2018skipnet, graves2016adaptive, figurnov2017spatially, mcgill2017deciding, meng2020ar}. BlockDrop \cite{blockdrop} dynamically selects layers per sample, while GaterNet \cite{geternet} learns channel-wise binary gates. Channel gating \cite{hua2019channel} skips computation on less impactful regions, and SpotTune \cite{spottune} adapts routing through fine-tuned or pre-trained layers. Adaptive region selection for fast object detection is explored in \cite{najibi2019autofocus, gao2018dynamic}. Inspired by \cite{wu2019adaframe, meng2020ar}, which use memory-augmented LSTMs for frame or modality selection \cite{adamml, wang2021adaptive}, we extend this strategy with cross-modal distillation in a unified framework.

\vspace{-0.75em}
\section{Our Approach: \ourapproach}

\customsubsection{Problem Formulation}


Given multisensory data streams from wearable devices over $K$ input modalities $\{\mathcal{M}_1, \mathcal{M}_2, \cdots, \mathcal{M}_K\}$, our goal is to learn the optimal trade-off between performance and efficiency. To achieve the goal, we introduce a joint distillation and policy learning framework (\cref{main_diagram}), 
containing two key modules: a cross-modal distillation module (CFD) $\Phi$ and a task-aware multisensory policy learning module (TeMPLe) $\Pi$, which we discuss in \cref{distillation} and \cref{policy_learning}, respectively. 
We discuss our three-stage joint training strategy of these two modules in \cref{combined_training}, illustrated in \cref{main_diagram}.


Under a typical egocentric audio-visual setting, the modalities can be multi-channel audio $\mathcal{A}$, vision $\mathcal{V}$, and behavioral data (head pose, gaze, \etc) $\mathcal{B}$. 
While the distillation module $\Phi$ distills knowledge from a powerful yet costly teacher model to a lightweight student model for efficiently processing each modality,
the policy module $\Pi$ is responsible for ensuring the optimal usage of the modalities under disposal by learning to choose the best possible configuration at a given instance, \ie combination of audio channels, salient visual frames, or turning on/off a certain input modality altogether. 
Our pipeline is designed to address a variety of multisensory egocentric perception tasks. In this work, we demonstrate its applicability by tackling three tasks: \textit{active speaker localization}, \textit{action recognition}, and \textit{behavior anticipation}.

For \textit{active speaker localization and behavior anticipation}, we adopt a saliency-driven modality switching mechanism through TeMPLe to dynamically switch between modalities, activating only those sensors that offer the highest trade-off between performance and efficiency for a task at a given time.
\begin{equation}
    \mathcal{M}^t_s = \operatorname*{argmax}_\mathcal{M} f_s\left(\{\mathcal{M}_1, \mathcal{M}_2, \cdots, \mathcal{M}_K\}\right)
\end{equation}
Here, $\mathcal{M}^t_s$ is the most salient modality at time $t$, and $f_s(\cdot)$ represents the saliency score of the modality.

For {action recognition}, we employ a saliency-based adaptive frame-selection strategy that is uniquely powered by audio previewing. Audio previewing involves analyzing short segments of audio data to detect cues that may indicate the regions of interest to detect an action. Unlike previous approaches \cite{egodistill}, which sample more than one frames from the video, we enforce a strict one-frame policy for an action via TeMPLe. Given a sequence of frames $s_t$ from a long untrimmed video, the most salient frames are detected by the following equation
\begin{equation}
    \{
   s_t
   |
   \exists_{s_{t_1 : t_2}} > \delta :\ 
   s_t \in \operatorname*{argmax}_{k \in s_{t_i:t_{i+w}}(s_t)}\mathcal{A}_{\Pi(t)}
    \}
\end{equation}

Here, $s_t$, are the selected frames based on $\mathcal{A}_{\Pi(t)}$ for each window $s_{t_i:t_{i+w}}$, where $w$ is the window length and $i \in [0, t-w]$. A frame is chosen if its value exceeds threshold $\delta$. We determine $\delta$ through empirical evaluation.

Unlike existing methods \cite{adamml}, we unify distillation and policy learning by reconciling two fundamentally different learning processes: one that transfers continuous, smooth knowledge from a teacher model (distillation) and another that makes discrete, reward-driven decisions (policy learning). \ourapproach integrates these through $\Phi$ and $\Pi$ by managing interference between the learning signals and addressing training instabilities.






\vspace{0.05in}

\customsubsection{Cross-Modal Feature Distillation}
\label{distillation}
\vspace{0.02in}




%


We first introduce the \underline{C}ross-Modal \underline{F}eature \underline{D}istillation (CFD) module, where we leverage compute-intensive teacher models $\Omega$ to distill knowledge into smaller student model $\Phi$. For a short video clip, a set of sparsely sampled frames $\mathcal{I}$ often already captures key \emph{semantic} information. Complementary to this, the audio $\mathcal{A}$ and behavioral information $\mathcal{B}$ capture \emph{contextual} information. Moreover, $\mathcal{A}$ and $\mathcal{B}$ are very efficient to process as compared to the expensive visual modality. Specifically, we obtain $\Phi$ that achieves $\Phi(\mathcal{I}, \mathcal{A}, \mathcal{B}) \approx \Omega(\mathcal{V})$ \cite{egodistill}. Such a lightweight model will be able to approximate the result of the heavy video model while being much more efficient. Our approach is agnostic to the specific video model $\Omega$. 


From the visual frames $\mathcal{I}$, associated audio $\mathcal{A}$ and behavioral signals $\mathcal{B}$ (for ASL, behavior anticipation), we extract image features $\mathbf{z}_{\mathcal{I}} = f_{\mathcal{I}}(\mathcal{I})$, audio features $\mathbf{z}_{\mathcal{A}} = f_{\mathcal{A}}(\mathcal{A})$ and behavioral features $\mathbf{z}_{\mathcal{B}} = f_{\mathcal{B}}(\mathcal{B})$ using lightweight feature encoders $f_{\mathcal{I}}$, $f_{\mathcal{A}}$ and $f_{\mathcal{B}}$  respectively.
Then, we fuse $\mathbf{z}_{\mathcal{I}}$, $\mathbf{z}_{\mathcal{A}}$, and $\mathbf{z}_{\mathcal{B}}$ with a fusion network $\xi$ to obtain the fused feature $\mathbf{z}_{\phi}  = \xi(\mathbf{z}_{\mathcal{I}}, \mathbf{z}_{\mathcal{A}}, \mathbf{z}_{\mathcal{B}})$. Finally, a fully-connected layer uses the fused feature to predict class logits $\Phi(\mathcal{I}, \mathcal{A}, \mathcal{B}) \in \mathbb{R}^C $. 


We train $\Phi$ with a combination of three losses, as follows. First, we train $\Phi$ to approximate the original visual feature $\mathbf{z}_{\mathcal{V}}$ from the teacher model $\Omega$:
\vspace{-0.05in}
\begin{equation}
    \mathcal{L}_1 = \sum_{(\mathbf{z}_{\Omega_i}, \mathbf{z}_{\phi_i})} \left\| \mathbf{z}_{\Omega_i} - \mathbf{z}_{\phi_i} \right\|_1.
\vspace{-0.1in}
    \end{equation}

Training with $\mathcal{L}_1$ alone does not replicate performance of $\Omega$. So, we train $\Phi$ with a knowledge distillation loss:
\begin{equation}
    \mathcal{L}_{\text{KD}} = \sum_{(\mathcal{V}_i, \mathcal{I}_i, \mathcal{A}_i, \mathcal{B}_i)}
    \mathcal{D}_{\text{KL}} (\sigma(\Omega(\mathcal{V}_i)/\tau), \sigma(\Phi (\mathcal{I}_i, \mathcal{A}_i, \mathcal{B}_i)/\tau)),
\vspace{-0.05in}
\end{equation}
where $(\mathcal{V}_i, \mathcal{I}_i, \mathcal{A}_i, \mathcal{B}_i)$ represents the $i$-th clip, $\mathcal{D}_{\text{KL}}$ measures KL-divergence between the class logits of $\Omega$ and $\Phi$, and $\tau$ is a temperature parameter. $\mathcal{L}_{\text{KD}}$ provides soft guidance through the heavy teacher model output to train a generalizable student model.

Finally, to further encourage preservation of task specific features, we also compute the prediction loss:
\begin{equation}
    \mathcal{L}_{\text{GT}} = \sum_{(\mathcal{I}_i, \mathcal{A}_i, \mathcal{B}_i)}
    \mathcal{L}_{\text{CE}}( c_{i},  \sigma(\Phi (\mathcal{I}_i, \mathcal{A}_i, \mathcal{B}_i))),
    \label{eq:gt}
\end{equation}
where $c_i$ is the ground-truth label. The final training loss:
\begin{equation}
    \mathcal{L}_\Phi = \alpha \mathcal{L}_{\text{KD}} + (1-\alpha) \mathcal{L}_{\text{GT}} + \beta \mathcal{L}_1,
    \label{eq:full}
\end{equation}
where $\alpha$ controls balance between knowledge distillation and task specific training \cite{hinton2015distilling}, and $\beta$ controls weight for feature space matching ($\alpha, \beta$ ablations in supplementary). 

Our cross-modal feature distillation module is inspired by previous distillation-based approaches~\cite{egodistill,listentolook,gan2019self,aytar2016soundnet,garcia2018modality, gupta2016cross}. However, our approach \textit{extends these methods by incorporating a more comprehensive sensory suite} (video frames, multi-channel audio, and behavioral data) under a multitask setting, and we also \textit{jointly train} this module with policy learning, as discussed below.


\vspace{0.05in}

\customsubsection{Task-Aware Multisensory Policy Learning}
\label{policy_learning}
\vspace{0.02in}


We introduce our \underline{T}ask-Awar\underline{e} \underline{M}ultisensory \underline{P}olicy \underline{L}earning N\underline{e}twork (TeMPLe), which is designed to learn dynamic actions for optimally selecting video frames, audio channels, and modalities. Given the diverse characteristics and available modalities across different egocentric tasks and datasets, we use a task-specific action space. 
For active speaker localization and behavior anticipation, we learn an optimal policy that chooses which modalities to use. Specifically for audio, the policy selects a subset of audio channels rather than using all available channels. For action recognition, we initially use audio to identify temporal regions of interest and subsequently learn to select the most informative video frame within those regions. 


\vspace{0.05in}

\customsubsubsection{Active Speaker Localization.}
\label{asl_policy}
We design a policy network that contains lightweight modality-specific feature extractors and an LSTM module to output a binary policy per segment for each modality to represent which modalities to keep \cite{adamml}. During training, the policy network is jointly trained with the task network using Gumbel-Softmax sampling \cite{jang2016categorical}. During inference, the policy network first decides the optimal combination of modalities to use for the given segment, which is subsequently forwarded to the task network to predict the outputs.

At the $t$-th time step, LSTM takes in joint feature $f_t$ of current segment $s_t$, previous hidden states $h_{t-1}$, cell outputs $o_{t-1}$ to compute current hidden $h_t$ and cell states $o_t$:
\vspace{-1.5mm}
\begin{equation}
    \label{eq:lstm}
    h_t, o_t = \text{LSTM}(f_t, h_{t-1}, o_{t-1}).
\vspace{-1mm}
\end{equation}
Given the hidden state, the policy network estimates a distribution for each modality and samples binary decisions $u_{t,k}$ that indicate whether to select modality $k$ at time step $t$ ($\mathbf{U} = \{\mathbf{u}_{t,k}\}_{l\le T, k\le K}$) via Gumbel-Softmax operation. Based on these decisions, at a given instance the corresponding modality inputs are forwarded to the next module.

\vspace{1mm}
\noindent{\textbf{Gumbel-Softmax Sampling.}} The policy thus learned is discrete thereby making the network training non-differentiable which hinders employing standard backpropagation. 
Inspired by \cite{adamml}, we adopt Gumbel-Softmax sampling \cite{jang2016categorical} to resolve this non-differentiability and enable direct optimization of the discrete policy in an efficient way.
Specifically, at each time step $t$, we first generate the logits $z_k\in\mathbb{R}^2$ (\ie, output scores of policy network for modality $k$) from hidden states $h_t$ by a fully-connected layer $z_k={FC}_k(h_t,{\theta_{FC}}_k)$ for each modality and then use the Gumbel-Max trick \cite{jang2016categorical} to draw discrete samples from a categorical distribution as: 
\vspace{-0.05in}
\begin{equation} 
\label{eq:gumbelmax}
    \hat{P}_k = \arg\max_{i\in \{0,1\}} (\log z_{i,k}+G_{i,k}), \ \ \ \ \ k \in [1, ..., K] \vspace{-1mm}
\end{equation} 
where $G_{i,k}=-\log(-\log U_{i,k})$ is a standard Gumbel distribution with $U_{i,k}$ sampled from a uniform i.i.d $Unif(0,1)$.
$\hat{P}_{k}$ is represented as a one-hot vector and is relaxed to a real-valued vector ${P}_k$ using softmax: 
\vspace{-1mm}
\begin{equation}
\label{eq:one}
    P_{i,k}=\frac{\exp((\log z_{i,k}+G_{i,k})/\tau)}{\sum_{j\in \{0,1\}} \exp((\log z_{j,k}+G_{j,k})/\tau)}, 
\vspace{-1mm}
\end{equation}
where $i\in \{0,1\}, \ k \in [1, ..., K]$, $\tau$ denotes temperature that controls the discreteness of $P_{k}$. Intuitively, when $\tau$ approaches 0 the samples from the Gumbel Softmax follows the discrete distribution.



\begin{table}[!t]
 \resizebox{\columnwidth}{!}{
\begin{tabu}{l|ccccc}
\hline
\multicolumn{1}{l|}{\textbf{Method}} & \begin{tabular}[c]{@{}c@{}}\textbf{Input} \\ \textbf{resolution} $\downarrow$ \end{tabular} & \textbf{Verb}$\uparrow$ & \textbf{Noun}$\uparrow$ & \textbf{Action}$\uparrow$ & \textbf{GMACs}$\downarrow$ \\ 

\hline

\rowfont{\color{lightgray}}
MoViNet-A6 \cite{kondratyuk2021movinets}                                                  & 320 $\times$ 320                      & 72.24          & 57.31          & 47.79            &            79.35    \\




\rowfont{\color{lightgray}}
TBN \cite{kazakos2019epic}                                                         & 224 $\times$ 224                      & 66.03          & 47.24          & 36.72            &         75.73       \\





\rowfont{\color{lightgray}}
AdaFuse \cite{adafuse}                                                     & 224 $\times$ 224                      & 65.52  & 55.75  & 50.16    &      95.84    \\

\rowfont{\color{lightgray}}M\&M      & 224 $\times$ 224            &    71.3                          &  66.2                           &          53.6                       &                  $98.8$             \\

\rowfont{\color{lightgray}}LaViLa     &  224 $\times$ 224           &   72.0                           &     62.9                         &           51.0                     &            $185$                   \\

\rowfont{\color{lightgray}}
Ego-only \cite{ego-only}                                                     & 224 $\times$ 224                      & 73.33  & 59.48  & 52.59    &   507.39       \\


\hline

ListenToLook \cite{listentolook}                                                     & 224 $\times$ 224                      & 61.27  & 52.52  & 39.85    &     380.46     \\

AdaMML \cite{adamml}                                                     & 224 $\times$ 224                      &  64.95 & 55.27  & 41.73    &  277.76        \\

VS-VIO \cite{yang2022efficient}                                                     & 224 $\times$ 224                      & 61.37  & 52.21  & 38.07    &   106.97       \\



\hline

TIM AV \cite{tim}                                                     & 224 $\times$ 224                      & \textbf{77.19}  & \textbf{67.22}  & \textbf{57.57}    & 26.62         \\

\hline

\ourapproach w/o TeMPLe                                                      & 224 $\times$ 224                      & 68.34  & 59.02  & 50.88    &  5.79        \\

\rowcolor[HTML]{e7fae6}
\textbf{
\ourapproach}                                             & \textbf{224 $\times$ 224}                      & \underline{76.65}         & \underline{66.83}         & \underline{56.74}           & \textbf{7.14}          \\ \hline
\end{tabu}
}
\vspace{-0.05in}
\caption{\textbf{Egocentric action recognition performance of baselines and other other SOTA on EPIC-Kitchens.
} We report the top-1 accuracy for verb, noun, and action (\%).}
\label{action_recognition_main_table}
\vspace{-0.05em}
\end{table}

\begin{table}[!t]
\centering
\resizebox{\columnwidth}{!}{
\begin{tabu}{l|cccc}
\hline
\textbf{Method} & \textbf{mAP}$\uparrow$ & \textbf{GMACs}$\downarrow$ & \textbf{Params (M)}$\downarrow$ & \textbf{Energy (J)}$\downarrow$ \\ \hline
\rowfont{\color{lightgray}}
MAVSL\textsubscript{\text{C+E}} \cite{jiang2022egocentric}                         & 86.32                                  & 6.852                                   &   16.13                                            &  0.698                                   \\





\rowfont{\color{lightgray}}
LocoNet \cite{wang2024loconet}                               &   71.83                               &   3.364                                 &   34.30                                            &       1.104                    
\\

\rowfont{\color{lightgray}}
Sync-TalkNet \cite{wuerkaixi2022rethinking}                               &  65.86                                &    3.788                                &     32.91                                          &   0.985                        
\\
\rowfont{\color{lightgray}}
ASD-Trans \cite{asd-transformer}                               &  70.13                                &    3.621                                &    15.03                                           &    0.482                       
\\
\rowfont{\color{lightgray}}
LW-ASD  \cite{liao2023light}                               & 71.60                                  &  1.280                                  &  5.36                                             &   0.145                        
\\

\hline

ListenToLook \cite{listentolook} & 71.28 & 12.452 & 17.34 & 1.032\\

AdaMML \cite{adamml} & 76.90 & 10.681 & 13.61 & 0.913 \\

VS-VIO \cite{yang2022efficient} & 72.31 & 7.873 & 5.97 & 0.266\\


\hline

MUST   \cite{swl}                               & \textbf{89.88}                                 &   0.642                                 &    2.17                                           &    0.029                       
\\ 

\hline

\ourapproach w/o TeMPLe & 78.59 & 0.077 & 0.36 & 0.003\\

\rowcolor[HTML]{e7fae6}
\textbf{\ourapproach}                                  &    \underline{89.74}                              &  \textbf{0.070}                                  &     \textbf{0.39}                                          &   \textbf{0.003}                                  \\ 
\hline
\end{tabu}
}
\vspace{-0.05in}

\caption{\textbf{Performance of active speaker localization on EasyCom.} We compare the mAPs (in \%) of various baselines in the visual field of view. Most of these tests use 4-channel audio. \ourapproach can dynamically choose optimal number of channels.}
\label{asl_main}
\vspace{-1em}
\end{table}

\begin{figure*}[!t]
  \begin{minipage}[b]{0.68\linewidth}
    \centering
    \renewcommand{\arraystretch}{1.05}
    \resizebox{\textwidth}{!}{
    \begin{tabular}{l|c|c|c|c|c|c|c|c|c|c}
    \hline
    \multirow{2}{*}{\textbf{Method}} & \multicolumn{3}{|c|}{Gaze} & \multicolumn{3}{|c|}{Orientation} & \multicolumn{3}{|c|}{Trajectory} & \multirow{2}{*}{Energy (J) $\downarrow$}\\
    \cline{2-10}
    & $T_{300 \mathrm{~ms}}$ & $T_{500 \mathrm{~ms}}$  & $T_{700 \mathrm{~ms}}$  & $T_{300 \mathrm{~ms}}$  & $T_{500 \mathrm{~ms}}$  & $T_{700 \mathrm{~ms}}$  & $T_{300 \mathrm{~ms}}$  & $T_{500 \mathrm{~ms}}$  & $T_{700 \mathrm{~ms}}$ & \\
     \hline
    \demph{MultitaskGP \cite{gardner2018gpytorch}} & \demph{11.42}  & \demph{15.59} & \demph{18.40} & \demph{\underline{4.70}} & \demph{9.28} & \demph{12.27} & \demph{13.75} & \demph{17.86} & \demph{20.02} & \demph{0.056} \\
     
     
    
    \demph{GazeMLE \cite{li2021eye}} & \demph{10.74}  & \demph{14.37} & \demph{18.14} & \demph{4.68} & \demph{9.11} & \demph{12.03} & \demph{14.33} & \demph{16.02} & \demph{18.64} & \demph{1.371}\\
    
    \demph{GLC \cite{lai2022eye}} & \demph{10.21}  & \demph{14.66} & \demph{17.80} & \demph{4.76} & \demph{8.98} & \demph{11.70} & \demph{13.15} & \demph{15.39} & \demph{17.41} & \demph{0.972}\\
    
    
    
    \hline
    
    ListenToLook \cite{listentolook} & 13.68 & 17.24 & 19.02 & 5.47 & 8.92 & 11.36 & 13.54 & 15.11 & 17.02 & 0.512\\
    
    AdaMML \cite{adamml} & 12.16 & 16.70  & 18.31 & 5.41  & 8.76 & 11.24 & 13.27 & 14.10  & 16.28 & 0.296\\
    
    VS-VIO \cite{yang2022efficient} & 14.83 & 19.27 & 20.54 & 8.41 & 12.44 & 13.19 & 15.71 & 16.92 & 18.53 & 0.097\\

    \hline
     
    $\mathrm{MuST}_{\mathcal{AVB}} \cite{swl}$ & \underline{9.17} & \underline{12.15} & 14.75 & 4.78 & \textbf{7.36} & \textbf{9.90} & 9.96 & 12.38 & 13.95 & 0.029\\

    \hline

    \ourapproach w/o TeMPLe & 10.95  & 14.69 & 16.18 & 5.20 & 7.88 & 10.81 & 11.50 & 13.66 & 12.98 & 0.003\\
     \rowcolor[HTML]{e7fae6}
    \textbf{\ourapproach} & \textbf{8.53} & \textbf{11.93} & \textbf{14.58} & \textbf{4.61} & \underline{7.39} & \underline{9.91} & \textbf{9.58} & \textbf{11.97} & \textbf{13.36} & \textbf{0.003} \\
    \hline
    \end{tabular}
    }
    \vspace{-0.05in}
    \captionof{table}{\textbf{Comparison of behavior anticipation errors on the AEA Dataset.} The energy values (in J) are reported by aggregating over three time windows ($T_{300 \mathrm{~ms}}$, $T_{500 \mathrm{~ms}}$, and $T_{700 \mathrm{~ms}}$).}
    \label{behaviour_aea_main}
    \end{minipage}   
    \hspace{0.01\textwidth}
    \begin{minipage}[b]{0.31\linewidth}
        \centering
        \renewcommand{\arraystretch}{1.28}
        \resizebox{\textwidth}{!}{
        \begin{tabular}{cc|cc|cc}
        \hline
        \multirow{2}{*}{\textbf{$\lambda_{K}$}} & \multirow{2}{*}{\textbf{$\gamma$}} &
        \multicolumn{2}{c|}{\textbf{AR}} & \multicolumn{2}{c}{\textbf{ASL}} \\
        \cline{3-6}
        & & \textbf{Acc.}$\uparrow$ & \textbf{GMACs}$\downarrow$
        & \textbf{mAP}$\uparrow$ & \textbf{GMACs}$\downarrow$ \\ \hline    [0, 0, 0]                               &   0                               &   56.99                                 &   13.68   &  89.77   & 0.391                                                                     \\
        
      [1, 1, 1]   & 1                                 & 56.27                                   &   9.23  &  89.48   & 0.102                                                                       \\
        
       [1, 0.05, 0.03]        & 1                                 & 56.83                                   &  7.67      &  89.76  &  0.092                                                  \\ 
        \rowcolor[HTML]{e7fae6}
        \textbf{[1, 0.05, 0.03]}                                & \textbf{10}                                 &   \underline{56.74}                                 &  \textbf{7.14}  & \underline{89.75}   & \textbf{0.070}                                                                           \\ \hline
        \end{tabular}
        }
        \captionof{table}{\textbf{Ablation on modality usage on AR and ASL.} We compare the performance and efficiency across modality usage on \ourapproach. Values of $\lambda_{K}$ are in the format [Visual, Audio, Behavior].}
        \label{modality usage results}
      \end{minipage}
\vspace{-1em}
\end{figure*}

\vspace{0.05in}

\customsubsubsection{Action Recognition.} 
Given a crop of untrimmed video $\mathcal{V}$ and its corresponding action label $\{\mbf v_i, \y_i\}_{i=1}^D$, we aim to perform action recognition by processing the minimum number of frames. We leverage $\mathcal{A}$ to help identify potential regions of interest. This module consists of three main components: 
\textit{(i)} an audio feature extraction module $g_{\mathcal{A}}(\mathcal{A})$, \textit{(ii)} a multi-head attention recurrent CNN handshaking module, and \textit{(iii)} LSTM layers.

\cref{main_diagram} illustrates that our audio guidance strategy determines three most salient regions (\textcolor{red}{red} dotted lines) from each of which one representative frames are chosen at $t$, $t+k_1$ and $t+k_2$ respectively. The extracted audio feature is processed by a sequence of multi-head attention $m_i$ and recurrent CNN layers $r_i$. We perform handshaking between the corresponding layers by learning learnable parameters $\rho_i$. We claim that although audio semantic information is captured by the multi-head attention layers, we need a separate time-aware module to encode temporal information. Given an intermediate latent audio feature, a single self-attention \cite{transformers} operation consist of $\bm{Q} = \bm{W}^q \bm{f}$, $\bm{K} = \bm{W}^k \bm{f}$, $\bm{V} = \bm{W}^v \bm{f}$: $\text{\textbf{\textit{Attention}}}(\bm{Q}, \bm{K}, \bm{V}) = \text{\textbf{\textit{Softmax}}}\left(\frac{\bm{Q} \bm{K}^T}{\sqrt{d_k}}\right)\bm{V}$, where $d_k$ is the dimension of the query and key features. To incorporate the additional guidance from recurrent CNN block (RCNN), the input to $i+1^{th}$ layer of RCNN $z_{l+1}^{MH}$ is:
\vspace{-0.05in}
\begin{equation}
    z_{l+1}^{RCNN} = z_{l}^{RCNN} + \rho z_l^{MH}
    \vspace{-0.03in}
\end{equation}

Finally, the LSTM layer ingests the temporal information aware audio features to detect the potentially distinct events $\mathcal{E}$ = $\{e_1, e_2, ..., e_n\}$. 
Given a long untrimmed audio input, we select one frame corresponding to each potential event $e_i$ where $i \in \{1, ..., n\}$ from say a partial sequence $s_{1:t}$, from the video based on the audio guidance:
\vspace{-0.05in}
\begin{equation}
    \s_{1:t} = [\s_{1:t-1}; \s_t], \quad \s_t = \mathcal{A}_{\Pi(t)},
    \vspace{-0.05in}
\end{equation}
where $\s_{1:t-1}$ denotes a partial clip of length $t-1$ and $\s_t$ is a single video frame. The frame level representation is then passed to the recognition module for subsequent processing.


\vspace{0.05in}

\customsubsubsection{Behavior Anticipation.}
We follow a similar set up in behavior anticipation task as ASL. Following \cite{swl} we consider three different directional behaviors of the wearer, i.e., eye gaze, head orientation, and motion trajectory which we collectively term as behavioral information $\mathcal{B}$. The action space for policy learning remains the same as \cref{asl_policy}.


\vspace{1mm}
\customsubsubsection{Policy Objective.}
During training, we minimize the following loss to encourage the optimal balance between performance and efficiency:
\vspace{-0.06in}
\begin{equation}
\label{eq:loss}
    \mathbb{E}_{(\mathcal{M},y)\sim \mathcal{D}_{train}}\left[-y\log(\mathcal{P}(\mathcal{M}; \Theta)) + \sum\limits_{k=1}^{K} \lambda_k \mathcal{C}_k\right]
\vspace{-0.06in}
\end{equation}
where $\mathcal{P}(\mathcal{M}; \Theta)$ and $y$ represent the prediction and ground truth label of the training multimodal input stream $\mathcal{M}$. This component is replaced by task-specific objectives from CFD module during stage 3 of training. Here, $\lambda_k$ represents the cost associated with processing $k$-th modality or segments. Similarly, $U_k$ represents the decision policy for $k$-th modality or segment and $\mathcal{C}_k=(\dfrac{|U_k|_0}{C})^2$ measures the fraction of segments that selected modality $k$ out of total $C$ video segments; when a correct prediction is produced. We penalize incorrect predictions with $\gamma$, which including $\lambda_k$ controls the trade-off between efficiency and accuracy. 

\begin{table}[!t]
\resizebox{0.48\textwidth}{!}{
\begin{tabular}{l|cll}
\hline
\multicolumn{1}{l|}{\textbf{Modality}} & \multicolumn{1}{c}{\begin{tabular}[c]{@{}c@{}}\textbf{Policy} \\ \textbf{driven?}\end{tabular}} & \multicolumn{1}{c}{\textbf{mAP}$\uparrow$} & \multicolumn{1}{c}{\textbf{GMACs}$\downarrow$} \\ \hline
Head Pose (HP) -only                       &               \textcolor{OrangeRed}{\ding{55}}                              &  45.36                                &   0.014                                 \\
Audio (1 channel)                      &               \textcolor{OrangeRed}{\ding{55}}                              &  64.78                                &  0.032                                  \\
Audio (4 channels)                      &               \textcolor{OrangeRed}{\ding{55}}                              &  81.27                                &  0.039                                  \\
Audio (4 channels) + HP                             &       \textcolor{OrangeRed}{\ding{55}}                                         &   83.22                               &  0.048                                  \\
Audio (4 channels) + Visual             &     \textcolor{OrangeRed}{\ding{55}}                                           &   83.44                               & 0.131                                   \\
Audio (4 channels) + Visual + HP        &       \textcolor{OrangeRed}{\ding{55}}                                         &  88.07                                &  0.148                                  \\

\hline
\rowcolor[HTML]{e7fae6}
\textbf{\ourapproach}                                   &      \textcolor{ForestGreen}{\ding{51}}                                       &     \textbf{89.74}                             &   \textbf{0.073}                                \\ \hline
\end{tabular}
}
\vspace{-0.05in}
\caption{\textbf{Effect of choice of modalities.} We systematically study the effect of various combinations of constituent modalities (visual, multi-channel audio, etc.) for ASL on the EasyCom dataset.}
\label{choice_of_modality}
\end{table}

\vspace{0.05in}

\begin{table}[!t]
\centering
\tiny
\resizebox{0.48\textwidth}{!}{
\begin{tabular}{l|ccc}
\hline
\textbf{Method} & \textbf{mAP}$\uparrow$ & \textbf{GMACs}$\downarrow$ & \textbf{ Params (M)}$\downarrow$ \\ \hline
CFD + random policy                                   &  67.41                                &   0.089                                 &   0.36                                                                              \\

CFD + Heuristics                                   &        65.88                          &     1.742                         & 0.36                                
\\
\hline
                     
Stage 1 + Stage 2                                 &  83.64                                &  0.073                                  &    0.38                                                            \\ 
\rowcolor[HTML]{e7fae6}
\textbf{\ourapproach (up to stage 3)}                                & \textbf{89.74}                                 & \textbf{0.073}                                   &   \underline{0.38}                                                                               \\ \hline
\end{tabular}
}
\caption{\textbf{Training stages ablation.} We ablate different training stages to compare their contributions on ASL. \ourapproach achieves best performance when joint training is employed.}
\label{training stage ablation}
\vspace{-1em}
\end{table}



\subsection{Combined Training}
\label{combined_training}
The final stage involves full finetuning of both $\Pi$ and $\Phi$ (\cref{main_diagram}).
Let $\Theta=\{\theta_{\Phi}, \theta_{LSTM},\theta_{{FC}_1},...,\theta_{{FC}_{K}},$ $\theta_{\chi_1},...,\theta_{\chi_{K}}\}$ denote the total trainable parameters in our framework, where $\theta_{\Phi}$: parameters in the distillation module, $\theta_{LSTM}$: parameters of LSTM used in the policy network respectively. $\theta_{{FC}_1},...,\theta_{{FC}_{K}}$: parameters of the FC layers that generate policy logits from the LSTM hidden states and $\theta_{\chi_1},...,\theta_{\chi_{K}}$ represent the parameters of $K$ sub-networks that are jointly trained for various AV tasks. Hence, the overall objective function is:
\vspace{-0.05in}
\begin{equation}
    \mathcal{L}_\Theta = \eta_1 \mathcal{L}_\Pi + \eta_2 \mathcal{L}_\Phi,
\label{eq:policy_distillation_loss}
    \vspace{-0.05in}
\end{equation}
where $\eta_1$, $\eta_2$ are weight factors of the policy and distillation modules respectively (ablations on $\eta$ are in supplementary). 


\section{Experiments}



\customsubsection{Datasets.}

\noindent{\textbf{EPIC-Kitchens \cite{epickitchens}}}: Contains 100 hours of egocentric videos capturing daily activities in kitchen environments. It consists of 89,977 segments of fine-grained actions. Following \cite{tim, omnivore} we directly predict the action out of 3,806 classes present in the train and validation set.

\noindent{\textbf{EasyCom \cite{easycom}}}: Consists of egocentric conversations for augmented reality applications, with a total of 380K video frames and their corresponding sensory inputs such as pose and multichannel audio.

\noindent{\textbf{Aria Everyday Activities (AEA) \cite{lv2024aria}}}: Contains diverse egocentric videos capturing daily activities, e.g., cooking, chatting comprising 143 recordings from 5 environments. 

\noindent{\textbf{Evaluation Metrics.}}
To measure action recognition performance, we report per-video top-1 accuracy on the validation set. For ASL task, we report mean Average Precision (mAP) values. While to evaluate localization performance, we report Mean Angular Errors (MAE) of behaviors.

\noindent{To benchmark efficiency, we measure computational cost with GMACs, Energy and \# parameters following \cite{egoexo4d} during inference. We defer further details to supplementary.}







\vspace{1mm}

\customsubsection{Baselines}


We compare our method against \cite{listentolook, swl, yang2022efficient}, which are equipped with multimodal policy learning or distillation. We also design a baseline by removing the policy learning module \ourpolicy from \ourapproach. We suitably adapt these for all three tasks to serve as baselines.

\noindent{\textbf{Listen to Look \cite{listentolook}}}: uses the audio-based feature distillation module originally for video action recognition.

\noindent{\textbf{AdaMML \cite{adamml}}}: introduces an adaptive multimodal learning framework with a dynamic model, designed to efficiently handle varying inputs in real-time applications.

\noindent{\textbf{VS-VIO \cite{yang2022efficient}}}: presents a framework that improves odometry performance by dynamically selecting relevant visual modalities in conjunction with inertial data.

\noindent{\textbf{\ourapproach w/o \ourpolicy}}: We eliminate the policy learning module and only evaluate against the CFD module.

\begin{figure*}[!t]
  \begin{minipage}[b]{0.32\linewidth}
    \centering
        
		\includegraphics[height=0.55\linewidth,width=\linewidth]{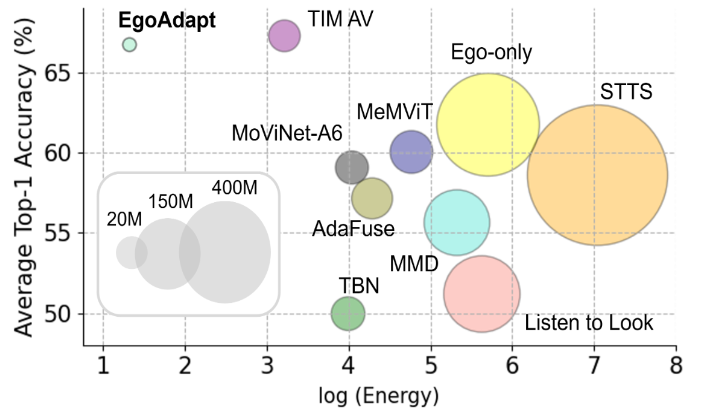}
        \vspace{-0.1in}
		\caption{\textbf{Acc. vs. log(Energy), size $\propto$ params for AR.} \ourapproach shows best-to-comparable performances with least compute when compared to SOTA methods.}
        \label{fig:efficiency_label_AR}
    \end{minipage}   
    \hspace{0.01\textwidth}
    \begin{minipage}[b]{0.32\linewidth}
    \centering
        
		\includegraphics[height=0.6\linewidth,width=\linewidth]{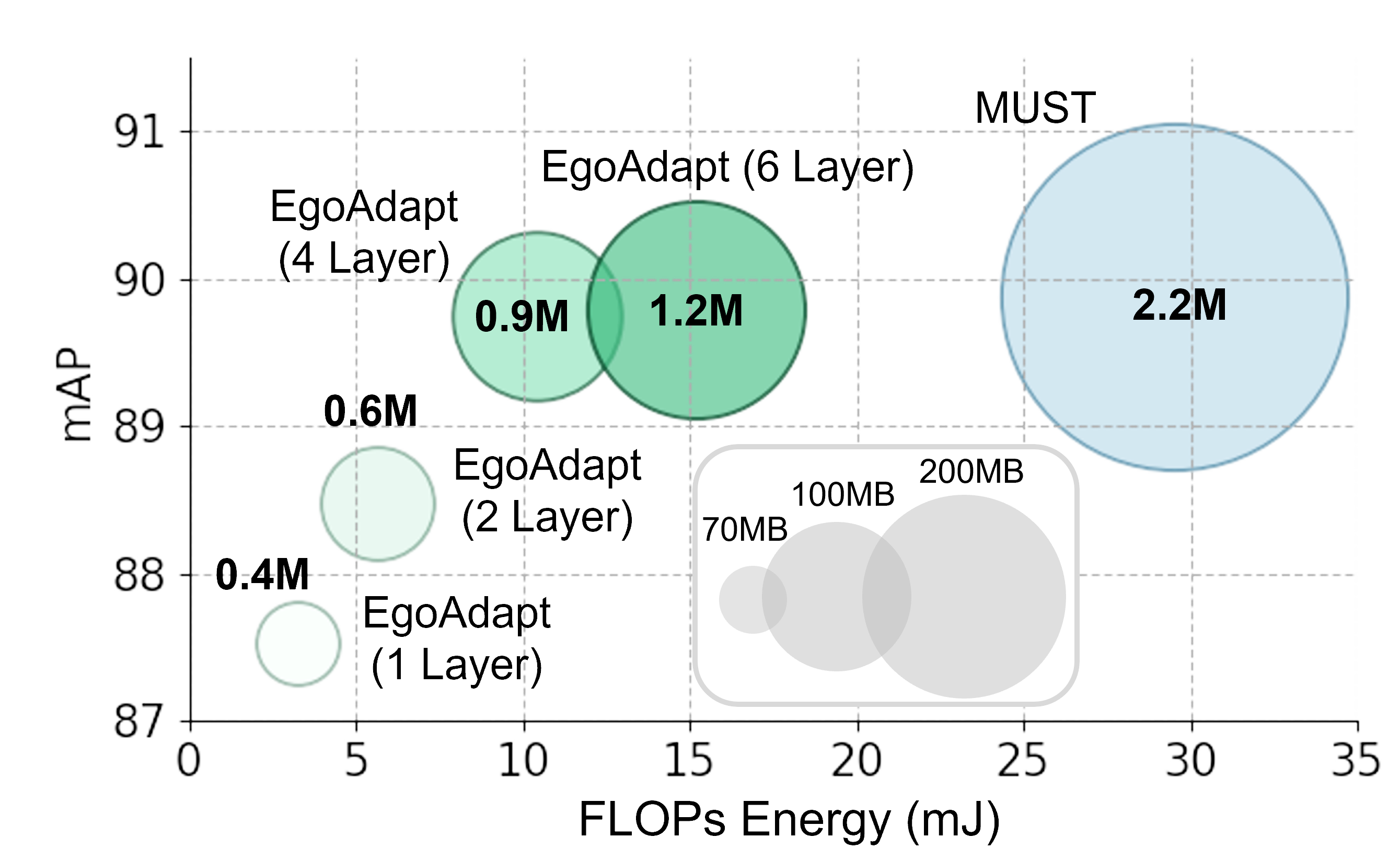}
\vspace{-0.1in}
		\caption{\textbf{mAP. vs. Energy for different variants of \ourapproach, size $\propto$ memory for ASL.} \ourapproach shows comparable performance even with 0.4M parameters.}
        \label{fig:efficiency_label_ASL}
    \end{minipage} 
    \hspace{0.01\textwidth}
    \begin{minipage}[b]{0.33\linewidth}
        \centering
        \renewcommand{\arraystretch}{1.8}
        \resizebox{\linewidth}{!}{
    
        \begin{tabular}{ccc|cc}
        \hline $\mathcal{L}_{\mathrm{KD}}$ & $\mathcal{L}_1$ & $\mathcal{L}_{\mathrm{GT}}$  &EK-100 (acc) $\uparrow$ & EC (mAP) $\uparrow$ \\ \hline
        
       \textcolor{OrangeRed}{\ding{55}} & \textcolor{ForestGreen}{\ding{51}} & \textcolor{ForestGreen}{\ding{51}}  & 50.06  & 80.54 \\
        
        \textcolor{ForestGreen}{\ding{51}} & \textcolor{OrangeRed}{\ding{55}} & \textcolor{ForestGreen}{\ding{51}} & 52.80  & 84.23 \\

        \rowcolor[HTML]{e7fae6}
        
        \textcolor{ForestGreen}{\ding{51}} & \textcolor{ForestGreen}{\ding{51}} & \textcolor{ForestGreen}{\ding{51}}  & \textbf{56.74}  & \textbf{89.74} \\


        
        \hline
        \end{tabular}
        }
        \vspace{-0.05in}

        \captionof{table}{\textbf{Ablation study of different loss components.} We compare the effect of different distillation and policy learning losses and report their performances on the corresponding datasets. EC: EasyCom, EK-100: EPIC-Kitchens-100.}
    \label{tab:loss_ablations}
      \end{minipage}
\end{figure*}

\begin{figure*}[t]
    \centering
    \includegraphics[width=\textwidth]{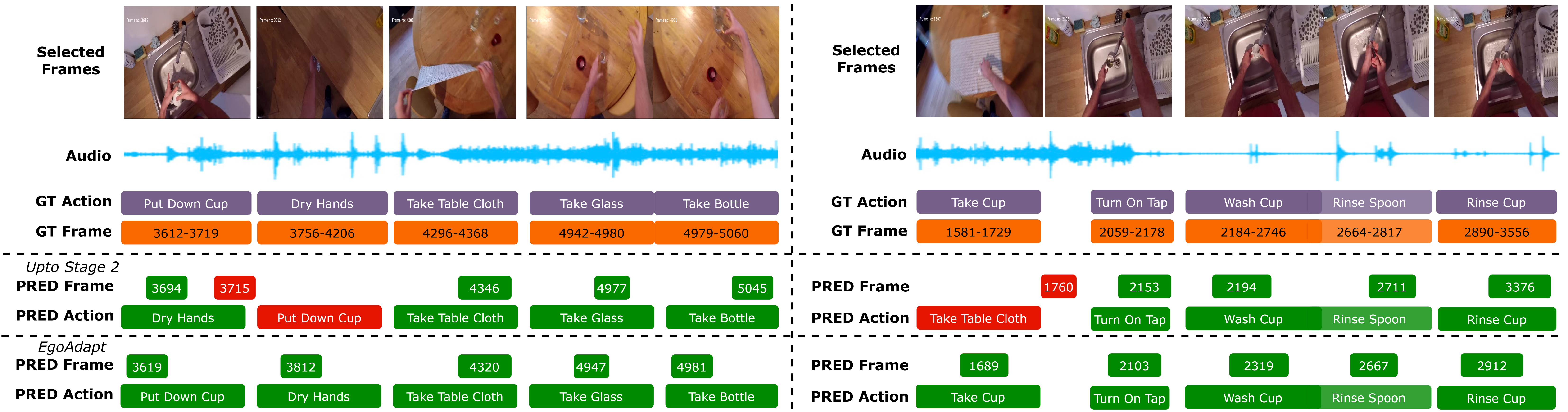}
    \vspace{-0.1in}
    \caption{\textbf{Qualitative examples of egocentric action recognition on EPIC-Kitchens.} The \textcolor{ForestGreen}{green} and \textcolor{red}{red} boxes represent correct and incorrect predictions, respectively. \ourapproach picks the most informative frame to predict the `Noun' classes, which is subsequently used to predict the action.}
    \label{fig:qual_ar_ek}
    \vspace{-2mm}
\end{figure*}




\noindent{\textbf{Task Specific Methods.}} We compare our method against existing ASL methods \cite{jiang2022egocentric} and adapt SOTA active speaker detection models \cite{liao2023light, wuerkaixi2022rethinking, wang2024loconet, asd-transformer}. For AR, we compare with \cite{kazakos2019epic, adafuse, kondratyuk2021movinets, ego-only}. For behavior anticipation, we adapt \cite{gardner2018gpytorch, li2021eye, lai2022eye} due to their strong performance on gaze-related tasks. More exhaustive comparison discussed in Supp.

\customsubsection{Main Results}


\noindent{\textbf{Comparison with State-of-the-Art Methods.}} \cref{action_recognition_main_table} compares \ourapproach with the SOTA models on EPIC-Kitchens. Our approach outperforms all baselines including the latest efficiency-driven approach Ego-only \cite{ego-only} by 3.35\% on verb, 7.84\% on noun and 4.98 \% on action respectively. We observe that our audio-visual model powered by the joint policy and distillation based training achieves a great balance between accuracy and efficiency by closely emulating teacher model's \cite{tim} performance.

We report the ASL performance of \ourapproach in \cref{asl_main}. Our approach operates at a significantly less compute budget (energy, parameters) yet achieves significant performance gains by outperforming efficiency focused approach LocoNet by 17.91\% on mAP score. Moreover, \ourapproach achieves similar performance as the teacher model (\cite{swl}) by operating at $\sim 82.02\%$ less parameters. 

Similarly, in \cref{behaviour_aea_main}, we note that \ourapproach consumes least energy (aggregated across 3 time windows) among all approaches and achieves best performances by reducing relative MAE score by 18.08\% over SOTA GLC \cite{lai2022eye}.




\begin{figure*}[t!]
    \centering
\includegraphics[width=0.98\textwidth]{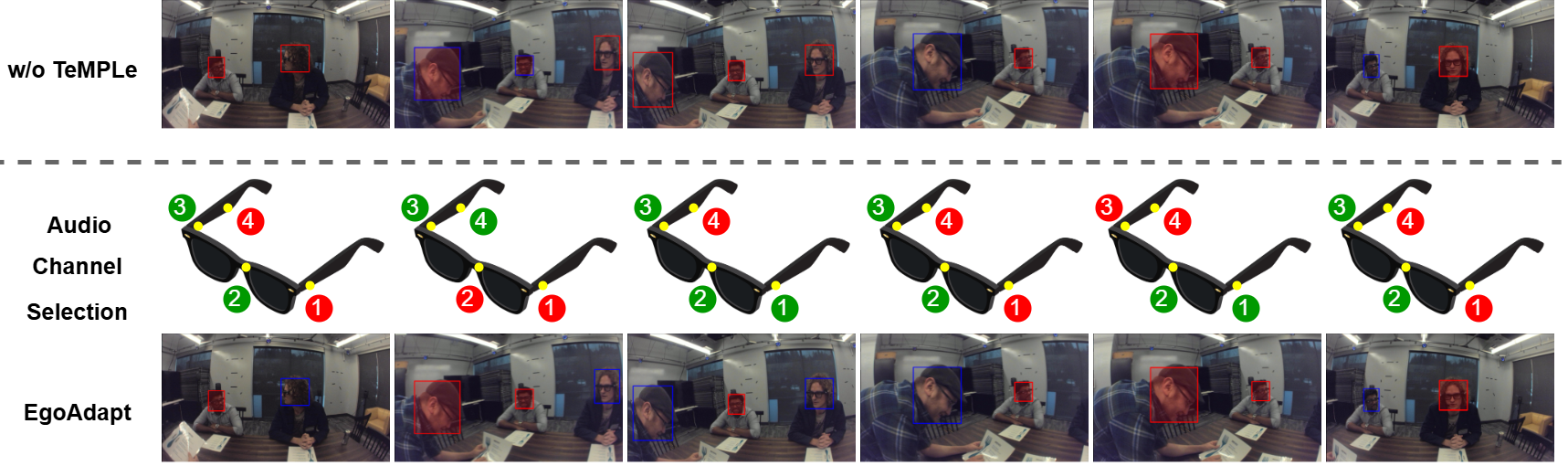}
    \caption{\textbf{Qualitative examples of ASL on EasyCom.}
    The \textcolor{red}{red}/\textcolor{blue}{blue} boxes indicate active/non-active speakers, and the red heatmap indicates model prediction. \ourapproach can make correct predictions for scenes with motion blur (col. 4), partial vision (col. 5), and multi-speakers (col. 2, 5) by performing intra-modal (audio channels) and inter-modal selections. The \textcolor{red}{red}/\textcolor{ForestGreen}{green} circles represent the discarded and selected audio channels.}
    \label{fig:qual_swl}
    \vspace{-2mm}
\end{figure*}


\customsubsection{On-Device Results}
We report the bit quantization results of the ASL task for both A and AV configurations in \cref{ondevice}. As seen, even the 4-bit model achieves encouraging performance operating at a very low power consumption. The proposed method runs in real-time at over 180 frames per second using a single GTX2080Ti GPU with about 28\% utilization. \ourapproach also has a smaller latency compared to traditional signal processing methods, which require estimating signal statistics over longer windows of time.

\begin{table}[!t]
\centering
\tiny
\resizebox{0.42\textwidth}{!}{
\begin{tabular}{c|cc|ccc}
\hline
\multicolumn{1}{c|}{\multirow{2}{*}{\begin{tabular}{@{}c@{}}
                   Precision \\
                   Level
              \end{tabular}}} & \multicolumn{2}{c|}{Modality}                & \multicolumn{1}{c}{\multirow{2}{*}{mAP $\uparrow$} } & \multicolumn{1}{c}{\multirow{2}{*}{Power (mW) $\downarrow$}} & \multicolumn{1}{c}{\multirow{2}{*}{\begin{tabular}{@{}c@{}}
                   Exec. \\
                   Time $\downarrow$
              \end{tabular}}} \\ \cline{2-3}
\multicolumn{1}{c|}{}                                  & \multicolumn{1}{c}{A} & \multicolumn{1}{c|}{V} & \multicolumn{1}{c}{}                     & \multicolumn{1}{c}{}                            \\ \hline
\multirow{2}{*}{4 bit}                                 &    \textcolor{ForestGreen}{\ding{51}}                   &       \textcolor{OrangeRed}{\ding{55}}                 & 77.14                                    & \textbf{7.38}               & \textbf{0.12}                             \\
                                                       &           \textcolor{ForestGreen}{\ding{51}}            &    \textcolor{ForestGreen}{\ding{51}}                    & 78.92                                    & 9.94      & 0.21                                      \\ 
                                                       
                                                       \hline
\multirow{2}{*}{8 bit}

  &         \textcolor{ForestGreen}{\ding{51}}              &        \textcolor{OrangeRed}{\ding{55}}                 & 80.56                                    & 11.37               & 0.33                            \\
                                                       &        \textcolor{ForestGreen}{\ding{51}}               &         \textcolor{ForestGreen}{\ding{51}}               & 81.13                                    & 14.90       & 0.42                                    \\ 
                                                       
                                                       \hline
\multirow{2}{*}{16 bit}

&     \textcolor{ForestGreen}{\ding{51}}                  &         \textcolor{OrangeRed}{\ding{55}}                & 84.39                                    & 19.11            & 0.59                               \\
                                                       &    \textcolor{ForestGreen}{\ding{51}}                   &         \textcolor{ForestGreen}{\ding{51}}               & 85.74                                    & 23.06                 & 0.68                          \\ 
                                                    
                                                    \hline
                                                       \multirow{2}{*}{32 bit}                                &     \textcolor{ForestGreen}{\ding{51}}                  &         \textcolor{OrangeRed}{\ding{55}}                & 83.22                                    & 29.71               & 0.89                            \\
                                                       &    \textcolor{ForestGreen}{\ding{51}}                   &         \textcolor{ForestGreen}{\ding{51}}               & \textbf{89.74}                                    & 34.39              & 1.00                                            \\ 
                                                       \hline
\end{tabular}
}
\caption{\textbf{Device realization results on ASL task.} Execution time values are normalized wrt the values of full precision runtime.}
\label{ondevice}
\vspace{-4em}
\end{table}


\customsubsection{Analysis}

\noindent{\textbf{Ablation on Modality Usage.}}
$\lambda$ and $\gamma$ controls the trade-off between accuracy and efficiency. Training without efficiency loss improves the performance slightly but results in much higher compute usage. Using equal cost weights for all modalities by setting $\lambda_i$=1 for $i \in \{1, ...,  K\}$ achieves a slight drop in performance with lesser utilization of cheaper modalities ($\mathcal{A}$, $\mathcal{B}$). Optimal performance is obtained with regulated weighting factors as shown in \cref{modality usage results}.

\noindent{\textbf{Choice of Modalities.}} \cref{choice_of_modality} compares performance of \ourapproach under various modality selection settings. While pose only method that has the least compute requirements, its performance is poor. \ourpolicy helps to learn the optimal balance between performance and efficiency.

\noindent{\textbf{Different Training Strategies.}} \cref{training stage ablation} shows that \ourapproach outperforms various training strategies and also achieves best efficiency results. The heuristic-based approach selects the downstream modalities based on a predefined scheme while random policy implies the modalities are selected randomly -- making \ourpolicy more effective than trivial strategies.


\noindent{\textbf{Ablation on Training Losses.}}
\cref{tab:loss_ablations} ablates different loss components. Training without $\mathcal{L}_{KD}$ leads to the largest performance drop, underlying importance of KD. Performance drop upon eliminating  $\mathcal{L}_{1}$ shows the importance of aligning features from the different modalities.

\noindent{\textbf{Efficiency Analysis.}}
\cref{fig:efficiency_label_AR} shows that \ourapproach achieves better average top-1 acc. for AR task when compared to SOTA model Ego-only \cite{ego-only} with $\sim$ 5x lesser energy in log scale (SOTA models are discussed in supplementary). \cref{fig:efficiency_label_ASL} compares \ourapproach across various model sizes and contrasts them with the teacher. Our final model operates at $\sim$ 5.5x less MACs while achieving almost similar performance -- indicating its superiority.

\vspace{0.05in}

\subsection{Qualitative Results}
\label{qualitative_results}
\vspace{-1mm}
\ourapproach can recognize actions across the two modalities including overlapping events (\cref{fig:qual_ar_ek}). We see consecutive actions are correctly recognized with varying interval lengths, such as the `Turn on Tap' and `Wash Cup' actions. The last section highlights the importance of the joint training stage, where our model improves its performance by combining CFD and \ourpolicy finetuning. 

\cref{fig:qual_swl} underlines the importance of intra-modal policy learning. \ourapproach, selects the salient audio channels to improve the performance compared to distillation-only module. Our model not only selects the optimal audio channels but also efficiently chooses other modalities such as $\mathcal{V}$ and $\mathcal{B}$ to inject spatial reasoning capability.

Similarly, in \cref{fig:qual_aea} the comparison between gaze prediction results indicates how correctly selecting the modalities can improve the future behavior prediction.

\begin{figure}[t!]
    \centering
\includegraphics[width=\columnwidth]{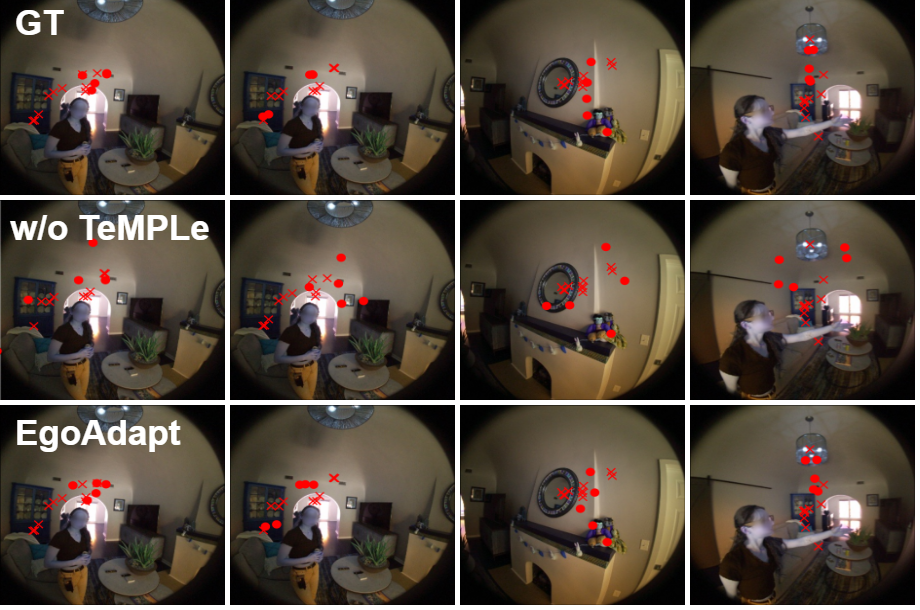}
    \caption{\textbf{Qualitative examples of egocentric behavior anticipation on the AEA Dataset.} Cross/circle symbols denote previous/anticipated behaviors. Our model reasonably anticipates future behaviors in common scenarios like long-term fixation and human-object interaction.}
    \label{fig:qual_aea}
    \vspace{-4mm}
\end{figure}





\vspace{-1.5mm}
\section{Conclusion and Future Work}
\vspace{-1.5mm}
We presented \ourapproach, a unified fully-differentiable framework that combines policy learning and cross-model distillation, which can effectively adapt to different egocentric perception tasks 
with the goal of achieving both competitive accuracy and efficiency. Experiments on benchmark datasets for AR, ASL, and BA demonstrate that our model significantly enhances efficiency while still on-par and, in many cases outperforming, the performance of state-of-the-art models. In the future, we plan to investigate how to further generalize our approach for other egocentric tasks such as hand-object interaction and long-term human activity understanding by incorporating more comprehensive action spaces such as optimal spatial resolution, network quantization, \etc. 

\medskip

{
    \small
    \bibliographystyle{ieeenat_fullname}
    \bibliography{main}
}

\newpage

\appendix


\clearpage

\twocolumn[
    \centering
    \Large
    \hspace{0.1em}%
    \noindent\rule{15cm}{0.6pt}
    \textsc{EgoAdapt}: Adaptive Multisensory Distillation and Policy Learning \\ for Efficient Egocentric Perception
    \\
    \vspace{0.5em}\textcolor{blue}{Supplementary Material} 
    \noindent\rule{15cm}{0.6pt}
    \vspace{1.0em}
]

\noindent{We add the following details in this supplementary:}

\noindent{\ref{supp_video} Supplementary Video} \\
\noindent{\ref{implementation_details} Model and Implementation Details} \\
\ref{more_related_works} More Related Works \\
\ref{selection_of_frames} Selection of Frames \\
\ref{other_teachers_comparison} Comparison with Other Teacher Models \\
\ref{sota_main_paper} Discussions on SOTA \\
\ref{efficieny_metrics} Details on Efficiency Metrics \\ 
\ref{noisy_condition} Performance in Noisy Condition\\
\ref{qualitative_examples} More Qualitative Examples \\
\ref{occlusion results} Occlusion and Conflicting signals results \\
\ref{failure_modes} Failure Modes \\
\ref{evaluation_metrics} Details on Evaluation Metrics \\
\ref{audio_channel_ablations} Ablation on Choice of Audio Channels \\
\ref{eta_ablations} Model Component Ablations \\
\ref{distillation_loss_ablations} CFD Loss Ablations \\
\ref{combined_algorithm} Combined Algorithm

\section{Supplementary Video}
\label{supp_video}
In our supplementary video, we have a brief overview of \ourapproach outlining all the training stages. We also add illustrative results for egocentric action recognition and active speaker localization tasks.

\section{Model and Implementation Details}
\label{implementation_details}

\noindent{\textbf{Active Speaker Localization \& Behaviour Anticipation.}}
The policy submodule is designed to efficiently process and integrate various data streams by using three lightweight modality encoders, one for video, one for audio, and one for sensor inputs. Each encoder has a compact architecture with $2$ convolutional layers, each followed by a max pooling operation and a ReLU activation. For video and audio encoder, we use $2-$D convolution and MaxPooling layer, whereas $1-$D  convolution and max pool are used for the sensor encoder. The first convolutional layer extracts essential low-level features from the input data, while the subsequent max pooling layer reduces spatial dimensions, thereby enhancing computational efficiency. The second convolutional layer further refines these features, and the ReLU activation introduces the non-linearity needed to capture complex patterns. This streamlined yet effective design enables the policy submodule to rapidly and accurately extract modality-specific features, ensuring that critical temporal and spatial information is retained for downstream decision-making processes regarding "\textit{which modality to use}." We use $K$ parallel FC layers on top of LSTM outputs to generate the binary decision policy for each modality or the chosen frame.

The Student module features $2$ multi-head attention (MHA) layers followed by a fully connected (FC) layer to efficiently capture and synthesize information. In the MHA module, we use $8$ heads.  The MHA layers allow the module to learn complex relationships by focusing on different parts of the input simultaneously, ensuring that various patterns and dependencies are recognized from multiple perspectives. Once these rich, context-aware representations are achieved, the FC layer of dimension $512$ integrates and consolidates the information into a final output, supporting decision-making or subsequent processing tasks. This design not only enhances the model's ability to capture nuanced patterns but also improves performance and efficiency.

\noindent{\textbf{Activity Recognition.}}
The policy submodule extracts time-aware audio features for action recognition by integrating three processing stages into one cohesive framework. Initially, $3$ multi-head attention layers with $8$ heads, enable the network to focus on different segments of the audio simultaneously, computing relationships between every pair of time steps and yielding refined features that capture global dependencies. These features are then complemented by RCNN layers, which combine convolutional operations to extract local acoustic patterns with recurrent processing to capture short-term dynamics. Each RCNN layer consists of a convolutional layer using a $3\times3$ kernel with $64$ filters, a stride of $1$, and \textit{“same”} padding to maintain spatial dimensions, followed by batch normalization and a ReLU activation. We use bidirectional LSTM with $128$ hidden units as the recurrent layer in RCNN. Information from both the MHA and RCNN layers is shared through a handshaking mechanism that ensures seamless integration of global and local insights. Finally, an LSTM layer of dimension $256$ synthesizes the combined long- and short-term features to determine the region of interest for action recognition accurately.

We follow the FasterNet \cite{chen2023run} architecture as our student model due to its capacity to reduce FLOPS and enhance efficiency. FasterNet is a hierarchical convolutional neural network comprising four stages that employs the Partial Convolution (PConv) operation to minimize computational redundancy and memory access. Each stage initiates with either an embedding or merging layer to downsample spatial dimensions and increase channel depth, followed by stacks of FasterNet blocks. Each FasterNet block first applies a $3\times3$ PConv that selectively processes the input channels, while the remaining channels bypass this operation. Next, a $1\times1$ pointwise convolution (PWConv) expands the channel dimension, and this is immediately followed by Batch Normalization and a ReLU activation function. This architecture, which resembles an inverted residual structure with shortcut connections, is designed to be straightforward, hardware-friendly, and efficient, making it well-suited for rapid inference within computational budgets.

\noindent{\textbf{Teacher Models.}}
We employ TIM \cite{tim} as the teacher model for the action recognition and MUST for \cite{swl} active speaker localization and behavior anticipation tasks, respectively.    

\noindent{\textbf{Hyper-parameters.}}
We set $\alpha = 0.90$ and $\beta = 0.85$ based on validation data. For EpicKitchens, we set $\tau=1.0$ and train for $150$ epochs. For EasyCom we set $\tau=10.0$ and train the model for $50$ epochs. 

\section{More Related Works}
\label{more_related_works}

\noindent{\textbf{Egocentric Video Understanding.}}
In the field of egocentric video understanding, numerous studies have demonstrated that incorporating additional modalities can greatly enhance performance \cite{zhang2022object, materzynska2020something, egovlpv2, kim2021motion, herzig2022object}. The hypothesis is straightforward: certain actions are more effectively understood through specific modalities. For instance, identifying that a person is `waving their hand' can be determined using motion trajectories alone \cite{materzynska2020something, radevski2021revisiting}. However, these studies assume that all modalities used during training are also accessible during inference and that the computational resources are sufficient to process modalities beyond the RGB frames. Consequently, some work \cite{wang2018videos, nagrani2021attention} have effectively used Faster-RCNN \cite{ren2016faster} and other object detection-specific models at inference time. In contrast, we argue that dynamically computing additional modalities for egocentric video understanding may be impractical. Thus, we propose a distillation and policy learning-based approach that learns how to optimally use various downstream modalities for efficient inference.  


\noindent{\textbf{Egocentric Action Recognition.}}
Incorporating contextual information, such as the motions of human body parts and details about active objects, offers a promising approach to egocentric action recognition. Several methods have been developed to leverage hand information, as the actor's hands provide crucial contextual cues \cite{tekin2019h+, kapidis2019multitask}. Some studies \cite{huang2020mutual, min2021integrating} have addressed action recognition by incorporating information about the actor's gaze, focusing on where they look during actions. Given that many actions in first-person videos involve interactions between the actor and objects, various methods have been developed to leverage information about the active objects \cite{fathi2013modeling, liu2017jointly}. Similar to our approach, several existing methods \cite{dessalene2021forecasting, li2015delving, zhou2016cascaded} integrate multiple types of contextual information for this task. While feature fusion is a promising strategy for boosting recognition performance, the associated increase in computational costs (e.g., larger model size) during inference is often overlooked. In contrast, our method learns to optimally use resources without compromising performance during inference.

\noindent{\textbf{Multi-modal Learning.}} 
Multimodal learning, in contrast to traditional single-modality approaches \cite{vdesirr, maw}, has made significant strides in areas such as cross-modal generation \cite{adverb, melfusion, foleygen, tang2024codi, magnet}, audio-visual representation learning \cite{listentopixel, audvisum, gao2024audio, sudarsanam2025representation}, multimodal large language models \cite{aurelia, meerkat, avtrustbench, vistallm}, and cross-modal integration \cite{apollo, intentometer, vlmnav, safari, volta, egovlpv2}. Recent studies have advanced cross-modal generation by utilizing visual and/or language context to produce coherent, complex audio \cite{adverb, melfusion}. Work on active audio-visual separation and embodied agents emphasizes the significance of motion and egocentric perception in developing robust representations. These concepts naturally extend to audio-visual LLMs \cite{vlmnav, avnav}, where perceptually grounded models engage with dynamic environments. In vision-language integration, recent research highlights the effectiveness of alignment across modalities \cite{apollo}. Collectively, these efforts illustrate the importance of dynamic, embodied perception in creating versatile multimodal systems.


\begin{table*}
\tiny
\centering
\resizebox{0.68\textwidth}{!}{
\begin{tabular}{l|cccc}
\hline
\multicolumn{1}{l|}{\textbf{Model}} & \multicolumn{1}{c}{\textbf{Verb}$\uparrow$} & \multicolumn{1}{c}{\textbf{Noun}$\uparrow$} & \multicolumn{1}{c}{\textbf{Action}$\uparrow$} & \multicolumn{1}{c}{\textbf{GMACs}$\downarrow$} \\ \hline
Fixed stride sampling & 72.32 & 59.25 &  46.81 & 7.14 \\
Random frame                 & 73.93                             & 60.86                             & 48.63                               & 7.14                              \\

First frame                  & 76.16                             & 61.63                             & 52.25                               & 7.14                              \\

Middle frame                 & 76.24                             & 64.06                             & 52.98                               & 7.14                              \\

All frames                   & \textbf{76.84}                             & \textbf{66.96}                             & \textbf{56.97}                               & \underline{16.51}                              \\
 \hline
\rowcolor[HTML]{e7fae6}
\textbf{\ourapproach}                     & \underline{76.65}                             & \underline{66.83}                             & \underline{56.74}                               & \textbf{7.14}                              \\ \hline
\end{tabular}
}
\caption{\textbf{Effect of visual frame selection strategy.} We compare the performance under different modes of key frame selection for action recognition task on Epic-Kitchens dataset.}
\label{frame_selection_results}
\end{table*}

\section{Selection of Frames}
\label{selection_of_frames}
\cref{frame_selection_results} compares the performance under different frame selection strategies. Upon randomly selecting a frame 
from potentially distinct activity regions (as determined by previewing audio) results in much inferior `Noun' detection performance which subsequently results in poor action recognition results. Similar performances are observed when the \emph{first} and \emph{middle} frames are chosen while there is a slight improvement in performance when all the underlying frames are chosen within a region of interest albeit at a much higher compute cost. We note, that \ourapproach can achieve a great balance between performance and efficiency by optimally choosing the most informative frame.

\section{Comparison with Other Teacher Models}
\label{other_teachers_comparison}

In this section, we compare the performance of \ourapproach while distilled from other teacher models for each task as outlined below.

\subsection{Action Recognition}
Similar to ASL, we employ MoViNet \cite{kondratyuk2021movinets}, MeMViT \cite{memvit}, MBT \cite{nagrani2021attention} as our teacher model as reported in \cref{supp_action_recognition}. The distilled student model coupled with \ourpolicy is able to closely emulate the teacher model's performance by operating at up to $\sim$ 28$\times$ less GMACs. 

\subsection{Active Speaker Localization}
We systematically employ TalkNet \cite{tao2021someone} and MAVASL \cite{jiang2022egocentric} in addition to MUST \cite{swl} as our teacher model to compare the performance of \ourapproach as reported in \cref{supp_asl}. Experimental results demonstrate that in all the cases our proposed approach is able to closely replicate the teacher model's performance while operating at a very low computation budget.

\subsection{Behavior Anticipation}
For behavior anticipation, the selected teacher models are SOTA approaches GazeMLE \cite{li2021eye} and GLC \cite{lai2022eye} (\cref{behaviour_aea_supp}). Experimental results indicate that \ourapproach is able to achieve similar performance as the heavy teachers (GMACs compared in the main paper) at a very lower compute cost.

The performance on these three tasks with varying teacher models underlines the efficiency of our proposed training paradigm. We claim that \ourapproach is teacher model agnostic and is able to replicate the heavy teacher model's performance while operating under a constrained compute setting thereby maintaining a good balance between performance and efficiency.

\begin{table*}
\centering
\tiny
 \resizebox{0.8\textwidth}{!}{
\begin{tabular}{l|ccccc}
\hline
\multicolumn{1}{l|}{\textbf{Method}} & \begin{tabular}[c]{@{}c@{}}\textbf{Input} \\ \textbf{resolution} $\downarrow$ \end{tabular} & \textbf{Verb}$\uparrow$ & \textbf{Noun}$\uparrow$ & \textbf{Action}$\uparrow$ & \textbf{GMACs}$\downarrow$ \\ 

\hline

MoViNet-A6 \cite{kondratyuk2021movinets}                                                  & 320 $\times$ 320                      & 72.24          & 57.31          & 47.79            &            79.35    \\

MeMViT \cite{wu2022memvit}                                                     & 224 $\times$ 224                      & 71.4          & 60.3          & 48.4            &       161.90         \\




MBT \cite{nagrani2021attention}                                                         & 224 $\times$ 224                      & 64.8          & 58.0          & 43.4            &     201.64           \\

TIM AV \cite{tim}                                                     & 224 $\times$ 224                      & \textbf{77.19}  & \textbf{67.22}  & \textbf{57.57}    & 26.62         \\

\hline


\ourapproach w/MoViNet & 224 $\times$ 224       &  71.46        &  56.32         &   46.14    &   7.14        \\


\ourapproach w/MeMViT & 224 $\times$ 224      &   70.67       &  59.91        &  48.25     &   7.14        \\

\ourapproach w/MBT & 224 $\times$  224     &    63.66      &   57.34       & 42.53      & 7.14          \\

\rowcolor[HTML]{e7fae6}
\textbf{\ourapproach w/TIM} & \textbf{224 $\times$ 224}      & \underline{76.65}         & \underline{66.83}         & \underline{56.74}      & \textbf{7.14}          \\ 

\hline
\end{tabular}
}
\vspace{-0.05in}
\caption{\textbf{Comparison with other models as teachers on EPIC-Kitchens.
} We report the top-1 accuracy for verb, noun, and action (\%).}
\label{supp_action_recognition}
\vspace{-0.05em}
\end{table*}

\begin{table*}
\centering
\tiny
\resizebox{0.8\textwidth}{!}{
\begin{tabular}{l|cccc}
\hline
\textbf{Method} & \textbf{mAP}$\uparrow$ & \textbf{GMACs}$\downarrow$ & \textbf{Params (M)}$\downarrow$ & \textbf{Energy (J)}$\downarrow$ \\ \hline

TalkNet \cite{tao2021someone}                              & 69.13                                 &   3.17                                 &      15.7                                         &   0.518                                  \\

MAVSL \cite{jiang2022egocentric}                         & 86.32                                  & 6.85                                   &   16.13                                            &  0.698                                   \\

MUST   \cite{swl}                               & \textbf{89.88}                                 &   0.642                                 &    2.17                                           &    0.029                       
\\ 

\hline


\ourapproach w/TalkNet                         &     68.90    & 0.070                                   &       0.39                                        &    0.003                                 \\

\ourapproach w/MAVASL                          &    85.74        &  0.070                                    &   0.39                                            &  0.003                                   \\

\rowcolor[HTML]{e7fae6}
\textbf{\ourapproach w/MUST}                                  &    \underline{89.74}                              &  \textbf{0.070}                                  &     \textbf{0.39}                                          &   \textbf{0.003}                                  \\ 
\hline
\end{tabular}
}
\vspace{-0.05in}

\caption{\textbf{Performance of active speaker localization on EasyCom with other teachers.}}
\label{supp_asl}
\vspace{-1em}
\end{table*}


\begin{table*}
    \centering
    \renewcommand{\arraystretch}{1.1}
    \resizebox{0.83\textwidth}{!}{
    \begin{tabular}{l|c|c|c|c|c|c|c|c|c}
    \hline
    \multirow{2}{*}{\textbf{Method}} & \multicolumn{3}{|c|}{Gaze} & \multicolumn{3}{|c|}{Orientation} & \multicolumn{3}{|c}{Trajectory} \\
    \cline{2-10}
    & $T_{300 \mathrm{~ms}} \downarrow$ & $T_{500 \mathrm{~ms}} \downarrow$  & $T_{700 \mathrm{~ms}} \downarrow$  & $T_{300 \mathrm{~ms}} \downarrow$  & $T_{500 \mathrm{~ms}} \downarrow$  & $T_{700 \mathrm{~ms}} \downarrow$  & $T_{300 \mathrm{~ms}} \downarrow$  & $T_{500 \mathrm{~ms}} \downarrow$  & $T_{700 \mathrm{~ms}} \downarrow$\\
     \hline
     
     
    
    GazeMLE \cite{li2021eye} & \demph{10.74}  & \demph{14.37} & \demph{18.14} & \demph{4.68} & \demph{9.11} & \demph{12.03} & \demph{14.33} & \demph{16.02} & \demph{18.64} \\
    
    \demph{GLC \cite{lai2022eye}} & \demph{10.21}  & \demph{14.66} & \demph{17.80} & \demph{4.76} & \demph{8.98} & \demph{11.70} & \demph{13.15} & \demph{15.39} & \demph{17.41} \\
    
    
    
    
    
    

     
    $\mathrm{MuST}_{\mathcal{AVB}} \cite{swl}$ & \underline{9.17} & \underline{12.15} & 14.75 & 4.78 & \textbf{7.36} & \textbf{9.90} & 9.96 & 12.38 & 13.95 \\

    \hline

    \ourapproach w/GazeMLE & 10.91  & 14.61 & 18.78 & 4.74 & 9.89 & 12.60 & 14.85 & 16.36 & 18.98 \\

    \ourapproach w/GLC & 10.87  & 14.76 & 19.32 & 5.10 & 9.24 & 12.09 & 13.72 & 15.80 & 17.94 \\
    
     
     \rowcolor[HTML]{e7fae6}
    \textbf{\ourapproach w/MUST} & \textbf{8.53} & \textbf{11.93} & \textbf{14.58} & \textbf{4.61} & \underline{7.39} & \underline{9.91} & \textbf{9.58} & \textbf{11.97} & \textbf{13.36}  \\
    \hline
    \end{tabular}
    }
    \vspace{-0.05in}
    \captionof{table}{\textbf{Comparison of behavior anticipation errors on the AEA Dataset with other teachers.}}
    \label{behaviour_aea_supp}
    \end{table*}

\section{Discussions on SOTA}
\label{sota_main_paper}
\subsection{Active Speaker Localization}
For a more comprehensive study, we compare our method against SOTA methods. MAVASL \cite{jiang2022egocentric} combine audio-only and audio-visual networks to perform spherical and inner field-of-view active speaker localization while LoCoNet \cite{wang2024loconet} learns a long-Short context network. Sync-TalkNet \cite{wuerkaixi2022rethinking} models cross-modal information with complex attention modules. ASD-Trans \cite{asd-transformer} employs a ResNet-18 to extract audio features. LW-ASD \cite{liao2023light} proposes a GRU based active speaker detection model.

\subsection{Action Recognition}
MoViNet \cite{kondratyuk2021movinets} proposes a three-step approach to improve computational efficiency while substantially reducing the peak memory usage of 3D
CNNs. TBN \cite{epic-fusion} introduces an audio-visual temporal binding network for egocentric action recognition. AdaFuse \cite{adafuse} presents an adaptive temporal fusion network for action recognition tasks. More recently, Ego-only \cite{ego-only} proposed an approach that enables action detection on egocentric videos without any form of exocentric (third-person) transferring.

\subsection{Egocentric Behavior Anticipation.}
Most existing works on egocentric gaze modeling target at egocentric gaze estimation rather than anticipation. We adapt gaze estimation models \cite{lai2022eye, li2021eye} to compare our method against them. We also report the performance of a competitive baseline of Multitask Gaussian Process \cite{gardner2018gpytorch}.

\section{Details on Efficiency Metrics}
\label{efficieny_metrics}

\subsection{GMACs}
We use the native PyTorch FLOP counter to get the total FLOP count in the forward pass. We convert this to GMACs (approximately 2 FLOPs = 1 MAC) by dividing with $10^9$.

\subsection{Energy Consumption} 
Accurately assessing the energy consumption of models is essential for their deployment in AR/VR devices \cite{abrash2021creating,chen2019eyeriss}. The energy expenditure arises from a complex interaction of factors, including sensors, computation, communication, data processing, memory transfers (SRAM and DRAM), and leakage—many of which are often overlooked when designing \emph{efficient} models, despite their significant contribution to overall energy use.

Following prior work \cite{sze2020evaluate, grauman2024ego}, we consider three key factors when modeling energy consumption: (1) The energy required for each model forward pass, determined by the number of operations (MACs). (2) The energy cost of memory read-write operations, including storing intermediate activations and model outputs. (3) The energy involved in activating, deactivating, and continuously operating sensors (e.g., camera, audio, IMU). 

We use GPUs as our processing device and employ the PyTorch memory profiler to capture a list of all operations performed during the forward pass (\texttt{model.forward()} call) along with their corresponding GPU memory usage. The total memory consumption is calculated as the sum of the memory costs for each individual operation.

For each modality, we track its active duration by counting the number of observations sampled that include the modality. We ensure that the sensors capture a minimum of 1 second's worth of samples.

\subsection{Trainable Parameters}
These are simply the learnable parameters that are adjusted during training to minimize the overall loss function.

\section{Performance in noisy conditions}
\label{noisy_condition}

\begin{figure}[t]
\centering\includegraphics[width=\columnwidth]{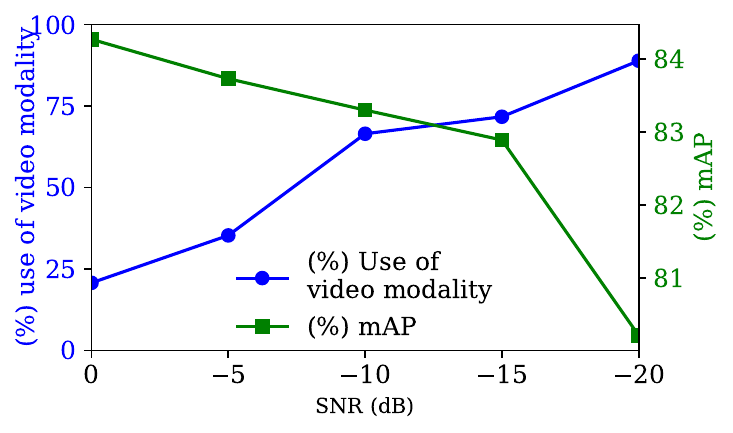}
  \vspace{-0.2in}
  \caption{\textbf{Adaptive Modality Utilization} of \ourapproach in the noisy scenario. \ourapproach increases video modality reliance to compensate for degraded audio, maintaining stable model performance under varying noise conditions.
  }
  
  \label{snr_v_perfromance}
  \vspace{-1em}
\end{figure}

\cref{snr_v_perfromance} illustrates the performance of the Active Speaker Localization (ASL) model under varying audio noise levels, quantified by signal-to-noise ratio (SNR). As the SNR decreases from -5 dB to -20 dB (indicating progressively noisier audio conditi`ons), the model adaptively increases its reliance on visual modalities, with video usage rising from 20.71\% to 88.89\%. Despite the degraded audio quality, the model maintains robust performance, evidenced by only a marginal decline in mAP from 84.27\% to 80.21\%. This underscores the effectiveness of the \ourapproach framework in dynamically leveraging complementary modalities through policy learning with the guidance of TeMPLe: under high noise, heightened video utilization compensates for unreliable audio cues, ensuring stable localization accuracy.

\section{More Qualitative Examples}
\label{qualitative_examples}
\begin{figure*}[h]
    \centering
\includegraphics[width=0.7\textwidth]{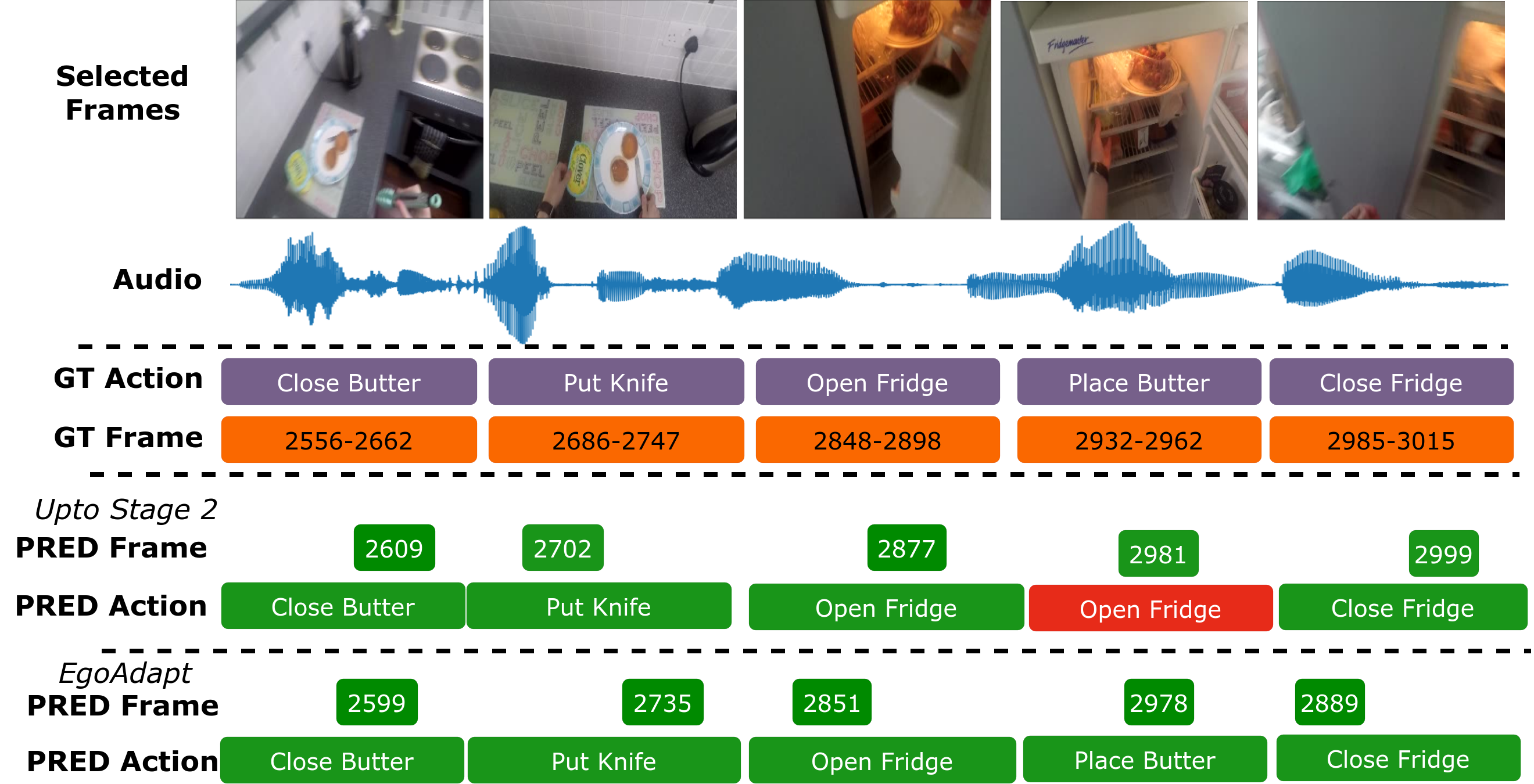}
    \caption{\textbf{More qualitative examples of Action Recognition on the Epic-Kitchens Dataset.}}
    \label{fig:supp_ek_qual}
\end{figure*}

\cref{fig:supp_ek_qual} compares action recognition performance between upto stage 2 training vs full training. As seen from the examples: the joint training of the two modules (CFD and \ourpolicy) indeed improves the performance by providing more accurate and tighter action prediction results. In the first and second example, after full finetuning, \ourapproach is able to choose a correct frame resulting in accurate action recognition.

\cref{fig:supp_ec_qual} compares the performance of \ourapproach with and without \ourpolicy for ASL task . We note that by learning to choose the useful combination of channels and sporadically making use of the visual modality our method is able to achieve improved performance when compared to without policy training.  

\cref{fig:qual_aea} illustrates more qualitative samples from egocentric behavior anticipation task. We contrast the performance with and without the policy learning module. As seen from the examples, after employing \ourpolicy the gaze prediction regions are more accurate and densely aligned to the GT.

\section{Occlusion and Conflicting signals results}
\label{occlusion results}
\begin{figure*}[h]
    \centering
    \includegraphics[width=\textwidth]{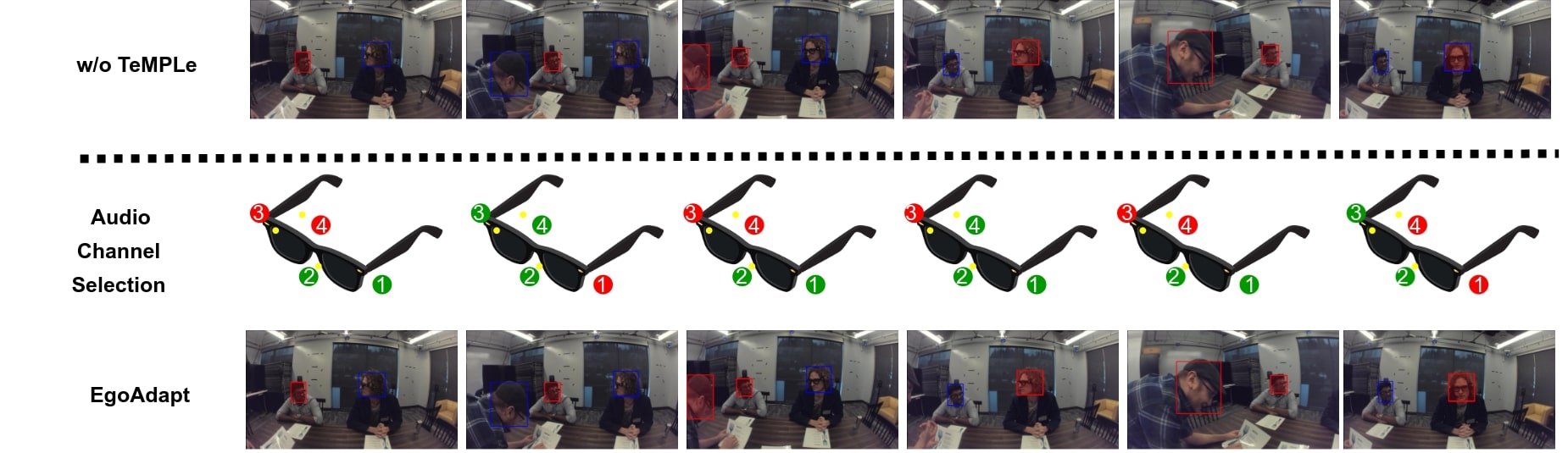}
    \caption{\textbf{More qualitative examples of Active Speaker Localization on the EasyCom Dataset.}}
    \label{fig:supp_ec_qual}
    \vspace{3mm}
\end{figure*}

Refer to \cref{fig:supp_ec_qual} -- last row column 3: \ourapproach can identify active speakers under heavy occlusion. The active speaker is looking away from the camera and his face is partially visible. As seen in the example, our approach is able to correctly identify the active speaker.   

In the same figure, the person is pretending to talk by taking notes. However, our model correctly relies on audio modality to eliminate such cases.

\section{Failure Modes}
\label{failure_modes}

\begin{figure*}[ht!]
    \centering
    \includegraphics[width=0.7\textwidth]{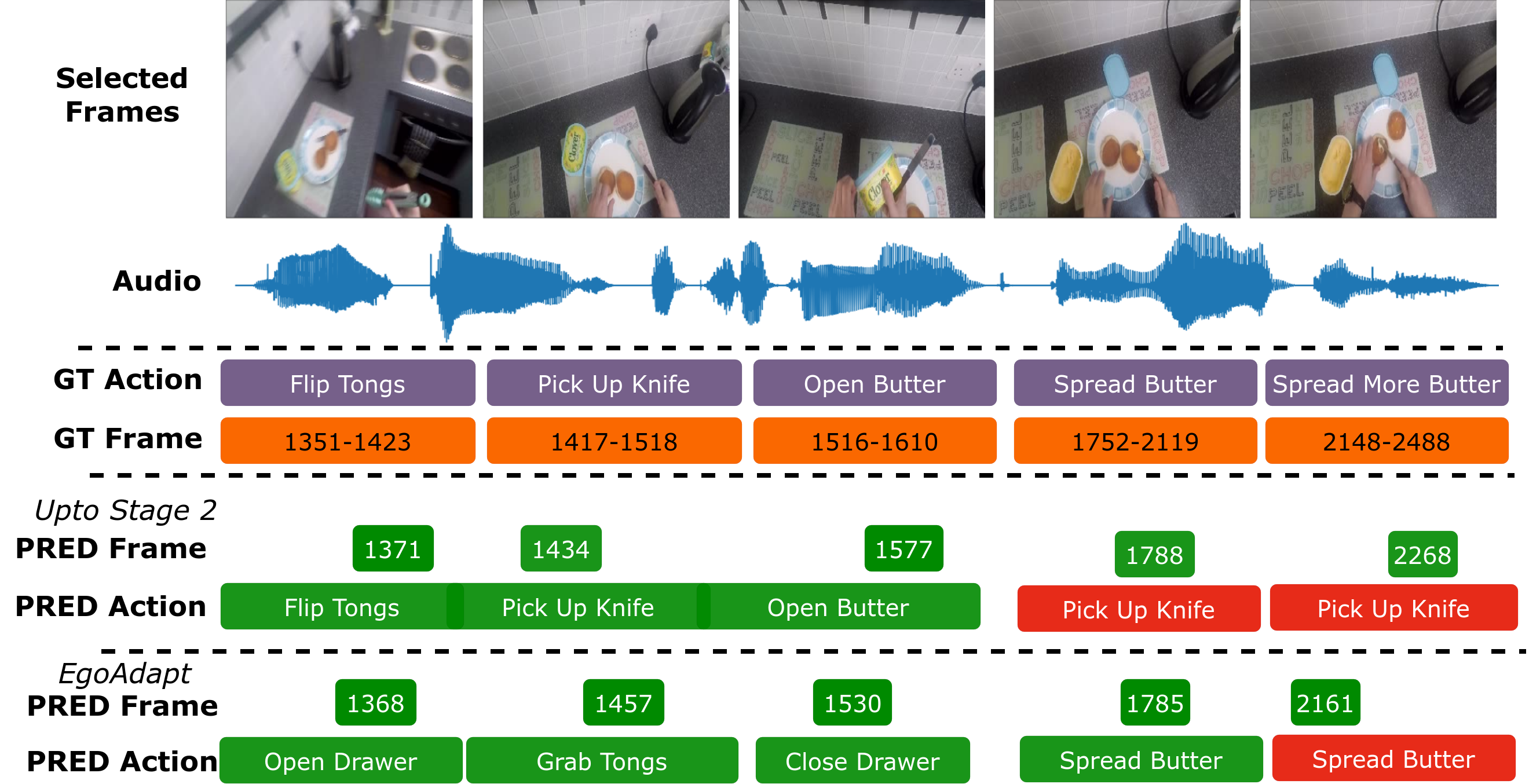}
    \caption{\textbf{Failure case for egocentric action recognition on EPIC-Kitchens dataset.}}
    \label{fig:failure_ek}
\end{figure*}

In this section we discuss failure modes of \ourapproach. In \cref{fig:failure_ek} we note that since our model is not equipped with fine grained understanding of the actions and not supervised by language models, it predicts \texttt{`spread butter'} while the original class label is \texttt{`spread more butter'}.

\begin{figure*}[ht!]
    \centering
\includegraphics[width=\textwidth]{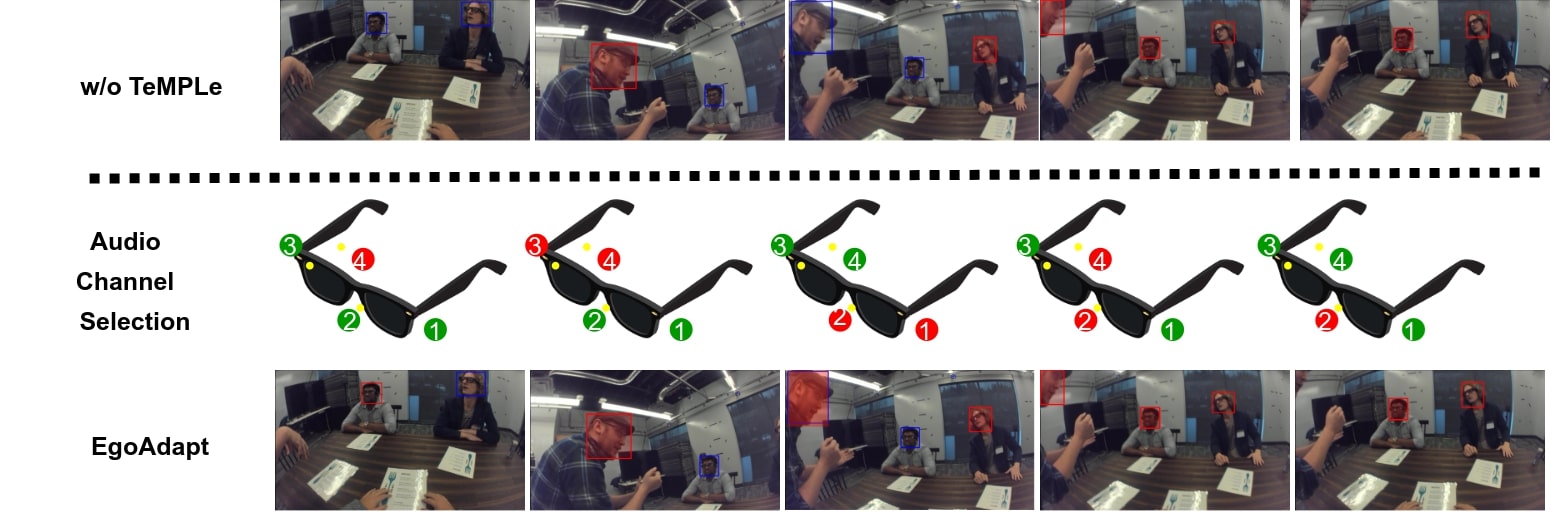}
    \caption{\textbf{Failure case for egocentric ASL on EasyCom Dataset.}}
    \label{fig:failure_ec}
\end{figure*}

Frame 3 in \cref{fig:failure_ec} represents a case where the person (with the hat) is actually non-active while holding his pose. \ourapproach due to its gaze bias is focusing towards this speaker resulting in wrong prediction results. While in frame 5, although the same speaker is active, our model is not able to detect him as he is partially visible.



\section{Details on Evaluation Metrics}
\label{evaluation_metrics}

\subsection{Action Recognition}

To measure action recognition performance, we report the per-video top-1 accuracy on the validation set. We densely sample clips from each video and average their predictions to compute accuracy.

\subsection{Active Speaker Localization}

We follow prior works’ \cite{swl, jiang2022egocentric} experimental settings for a fair comparison and report the mean average precision (mAP), which captures both spatial and temporal localization of speech activity inside the camera’s FOV. The mAP scores of all models are computed by pooling the maximum logit value within the corresponding head bounding boxes.

\subsection{Behavior Anticipation}
Our goal is to anticipate future behaviors in various reaction times (300/500/700ms) given the current audio-visual observations and previous behavioral contexts. To evaluate localization performance, we use Mean Angular Errors (MAE) of behaviors by comparing the argmax coordinate of the model’s prediction with ground truth
behaviors following \cite{swl}. MAE between prediction to ground truth reflects how far the model’s prediction deviates from the ground truth source direction on a sphere.

\section{Ablation on Choice of Audio Channels}
\label{audio_channel_ablations}
We compare the performance of \ourapproach under different predefined choices of audio channels in \cref{choice_of_audio_channels}. Although selecting all 4 channels results in strong ASL performance, the best performance is achieved when strategically all the downstream modalities are leveraged. As we employ very lightweight modules for each modality, the overall GMAC values remain within considerable limits.

\begin{table}[!t]
\resizebox{0.48\textwidth}{!}{
\begin{tabular}{l|cll}
\hline
\multicolumn{1}{l|}{\textbf{Audio Channels}} & \multicolumn{1}{c}{\begin{tabular}[c]{@{}c@{}}\textbf{Policy} \\ \textbf{driven?}\end{tabular}} & \multicolumn{1}{c}{\textbf{mAP}$\uparrow$} & \multicolumn{1}{c}{\textbf{GMACs}$\downarrow$} \\ \hline

[1, 2]                       &               \textcolor{OrangeRed}{\ding{55}}                              &    71.66                             &   0.059                                 \\

[1, 3]                      &               \textcolor{OrangeRed}{\ding{55}}                              &     73.38                             &    0.059                                \\

[1, 4]                      &               \textcolor{OrangeRed}{\ding{55}}                              &     68.25                             &  0.059                                  \\

[2, 3]                             &       \textcolor{OrangeRed}{\ding{55}}                                         &   69.57                               &    0.059                     \\

[2, 4]             &     \textcolor{OrangeRed}{\ding{55}}                                           &   66.19                               &  0.059                                  \\

[3, 4]        &       \textcolor{OrangeRed}{\ding{55}}                                         &     51.87                             &      0.059                              \\

[1, 2, 3]        &       \textcolor{OrangeRed}{\ding{55}}                                         &    82.69                              &    0.062                                \\ 

[1, 2, 4]        &       \textcolor{OrangeRed}{\ding{55}}                                         &    83.62                              &     0.062                               \\

[1, 3, 4]        &       \textcolor{OrangeRed}{\ding{55}}                                         &   80.48                               &    0.062                                \\

[2, 3, 4]        &       \textcolor{OrangeRed}{\ding{55}}                                         &    80.22                              &     0.062                               \\

[1, 2, 3, 4]        &       \textcolor{OrangeRed}{\ding{55}}                                         &     87.91                             &    0.068                                \\

\hline
\rowcolor[HTML]{e7fae6}
\textbf{\ourapproach}                                   &      \textcolor{ForestGreen}{\ding{51}}                                       &     \textbf{89.74}                             &   0.073                                \\ \hline
\end{tabular}
}
\vspace{-0.05in}
\caption{\textbf{Effect of choice of audio channels.} We systematically study the effect of various combinations of audio channels for ASL tasks on the EasyCom dataset.}
\label{choice_of_audio_channels}
\end{table}

\begin{algorithm*}[!t]
\small
\caption{\ourapproach: Training}
\label{algo:training}
\begin{algorithmic}[1]
\Require{Video/Frames: $\mathcal{V}$; Single/Multi-Channel Audio: $\mathcal{A}$; Behavior (IMU, Gaze) info: $\mathcal{B}$; 
Pre-trained Teacher: $\Theta_{\Omega}$; Cross-modal Feature Distillation Module: \textbf{CFD}; Policy Module: \textbf{TeMPLe}; Ground Truth Labels: GT; Task Type: $\mathcal{T}$; Loss Hyperparameters: $\eta_{1}, \eta_{2}$;}
\Ensure{Trained Policy (combination of sub-networks): $\Theta_{\Pi} = \{\Theta_{\text{LSTM}}, \Theta_{\text{FC}_{i}}, \Theta_{\chi_{i}}\}, \; \forall i \in \{1, 2, ..., k\}$; Trained Student: $\Theta_{\Phi}$.}
\State $\Theta_{\Phi} \leftarrow \text{\textbf{CFD}}(\Theta_{\Omega}, \mathcal{V}, \mathcal{A}, \mathcal{B})$ \Comment{\textit{Distillation (Stage 1.) using Eq. (4)}}
\While{not converged}
    \State $\Theta_{\Pi} \leftarrow \text{\textbf{TeMPLe}}(\Theta_{\Phi}, \mathcal{V}, \mathcal{A}, \mathcal{B}, \mathcal{T}, GT)$ \Comment{\textit{Policy Training (Stage 2) using Eq. (10).}}
    \State $\mathcal{L}_{\Theta} \leftarrow \eta_{1} \mathcal{L}_{\Pi} + \eta_{2} \mathcal{L}_{\Phi}$ \Comment{\textit{Combined Training (Stage 3) using Eq. (11).}}
    \State Optimize for $\Theta$ to reduce $\mathcal{L}_{\Theta}$ until convergence.
\EndWhile
\State \Return $\Theta = \{\Theta_{\Phi}, \Theta_{\Pi}\}$.
\end{algorithmic}
\end{algorithm*}

\section{Model Components Ablations}
\label{eta_ablations}

In \cref{eta_comparison} we compare the contributions of the two main model components, i.e. \ourpolicy ($\Pi$) and CFD ($\Phi$). We observe that lower values of $\eta_1$ and $\eta_2$ imply lesser weightage on both the components resulting in suboptimal performance. As we gradually add more weightage to the policy component the peeformance improves, however, since the task module is learned by CFD, higher values of $\eta_2$ boosts the overall performance for both the tasks.

\begin{table}[h]
\tiny
\centering
        \renewcommand{\arraystretch}{1.3}
        \resizebox{0.85\columnwidth}{!}{
    
        \begin{tabular}{cc|cc}
        \hline \bf $\eta_1$ & \bf $\eta_2$  & \bf EK-100 (acc) $\uparrow$ & \bf EC (mAP) $\uparrow$ \\ \hline
        
         0.10 & 0.20 & 45.22 & 68.19 \\
        
         0.50 & 0.20 & 46.21  & 70.35 \\

         0.75 & 0.20 & 47.80  & 71.48 \\

         0.95 & 0.20 & 49.52  & 72.81 \\

         0.95 & 0.50 &  52.68 & 75.95 \\

         0.95 & 0.80 & 56.08  & 88.28 \\

        \rowcolor[HTML]{e7fae6}
        
         \bf 0.95 & \bf 1.2  & \textbf{56.74}  & \textbf{89.74} \\


        
        \hline
        \end{tabular}
        }
\caption{\textbf{$\eta_1$ and $\eta_2$ ablation results.} }
\label{eta_comparison}
\end{table}

\section{CFD Loss Ablations}
\label{distillation_loss_ablations}
We ablate $\alpha$, $\beta$ in \cref{supp_distillation_ablations} to compare the contributions of the loss components in the distillation module. While $\alpha$ directly controls the relative weightage for the $\mathcal{L}_{KD}$;  $(1-\alpha)$ controls the weightage for the $\mathcal{L}_{GT}$ and $\beta$ controls that of $\mathcal{L}_1$ loss. We note that when $\alpha$ is small due to less weightage on $\mathcal{L}_{KD}$ the student model is not able to replicate the teacher model's performance resulting in inferior performance. Higher values of $\alpha$ improve performance. Similarly, as $\mathcal{L}_1$ helps to bring the feature level representations between the \emph{teacher} and \emph{student} models closer, higher values of $\beta$ result in better performance. However, we observe that the optimal results are obtained when $\alpha$ and $\beta$ are set to 0.90 and 0.85 respectively which helps to maintain a good balance between different loss components.

\begin{table}[h]
\tiny
\centering
        \renewcommand{\arraystretch}{1.3}
        \resizebox{0.85\columnwidth}{!}{
    
        \begin{tabular}{cc|cc}
        \hline \bf $\alpha$ & \bf $\beta$  & \bf EK-100 (acc) $\uparrow$ & \bf EC (mAP) $\uparrow$ \\ \hline
        
      0.10  & 0.50  & 43.63  & 78.09 \\
        
      0.40    & 0.50 & 46.15  & 80.23 \\

      0.70    & 0.50 & 48.73  & 82.49 \\

      0.90    & 0.50 & 51.41  & 85.57 \\

      0.90    & 0.75 &  53.54 & 87.80 \\

    \rowcolor[HTML]{e7fae6}
      \bf 0.90    & \bf 0.85 & \bf 56.74  & \bf 89.74 \\



        
        \hline
        \end{tabular}
        }
\caption{\textbf{$\alpha$ and $\beta$ ablation results.} }
\label{supp_distillation_ablations}
\end{table}


\section{Combined Algorithm}
\label{combined_algorithm}

We provide the overall \ourapproach training algorithm in \cref{algo:training}. Our key novelty is jointly training a policy and a distillation modules detailed in Line 3-4. Our approach can learn task-specific action spaces and coupled with distilled student modules achieve a great balance between performance and efficiency as shown through extensive experiments. 

\begin{figure*}[h]
    \centering
    \includegraphics[width=0.7\textwidth]{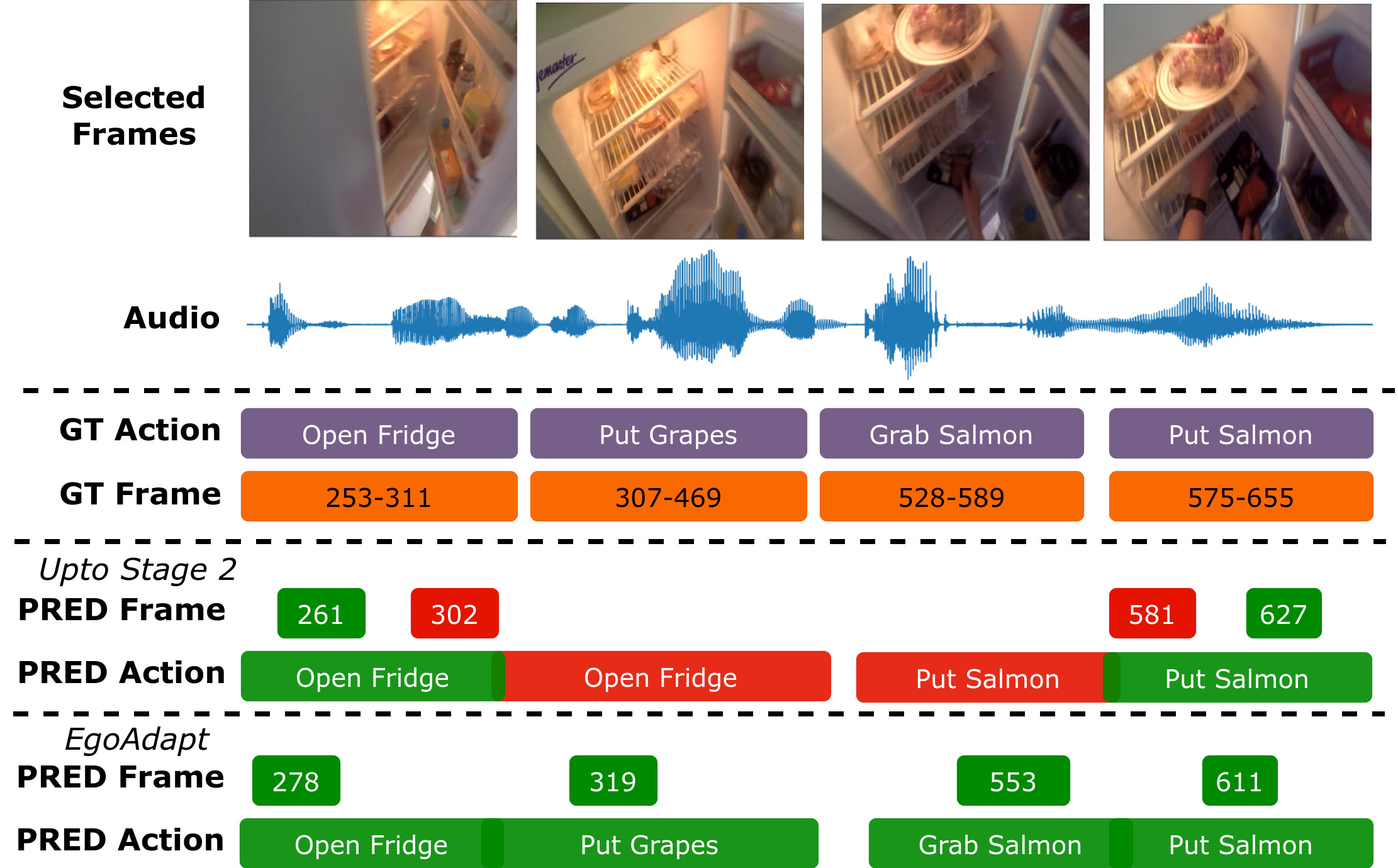}
    \vspace{3mm}
\end{figure*}

\begin{figure*}[h]
    \centering
    \includegraphics[width=0.7\textwidth]{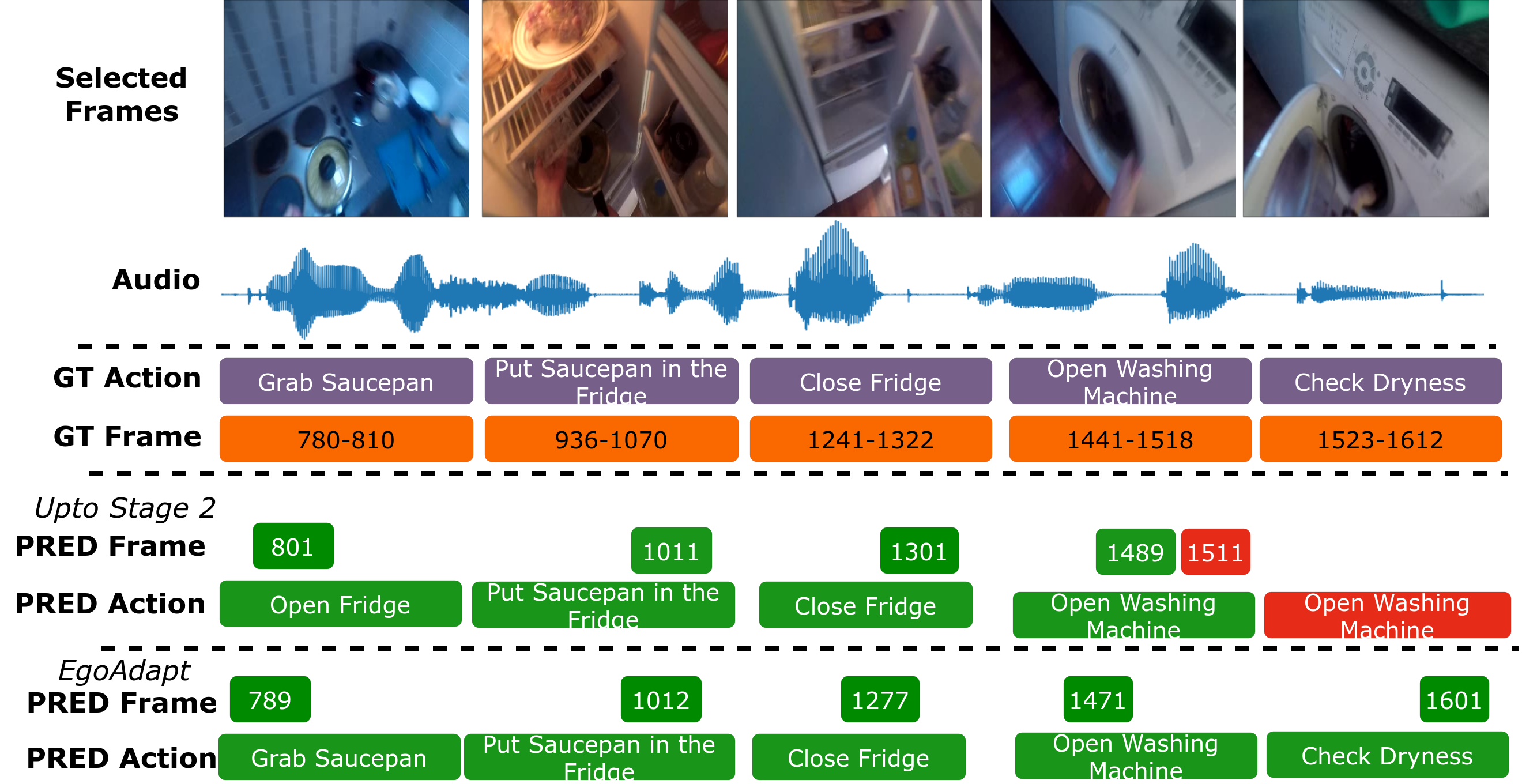}
    \vspace{3mm}
\end{figure*}

\begin{figure*}[h]
    \centering
    \includegraphics[width=0.7\textwidth]{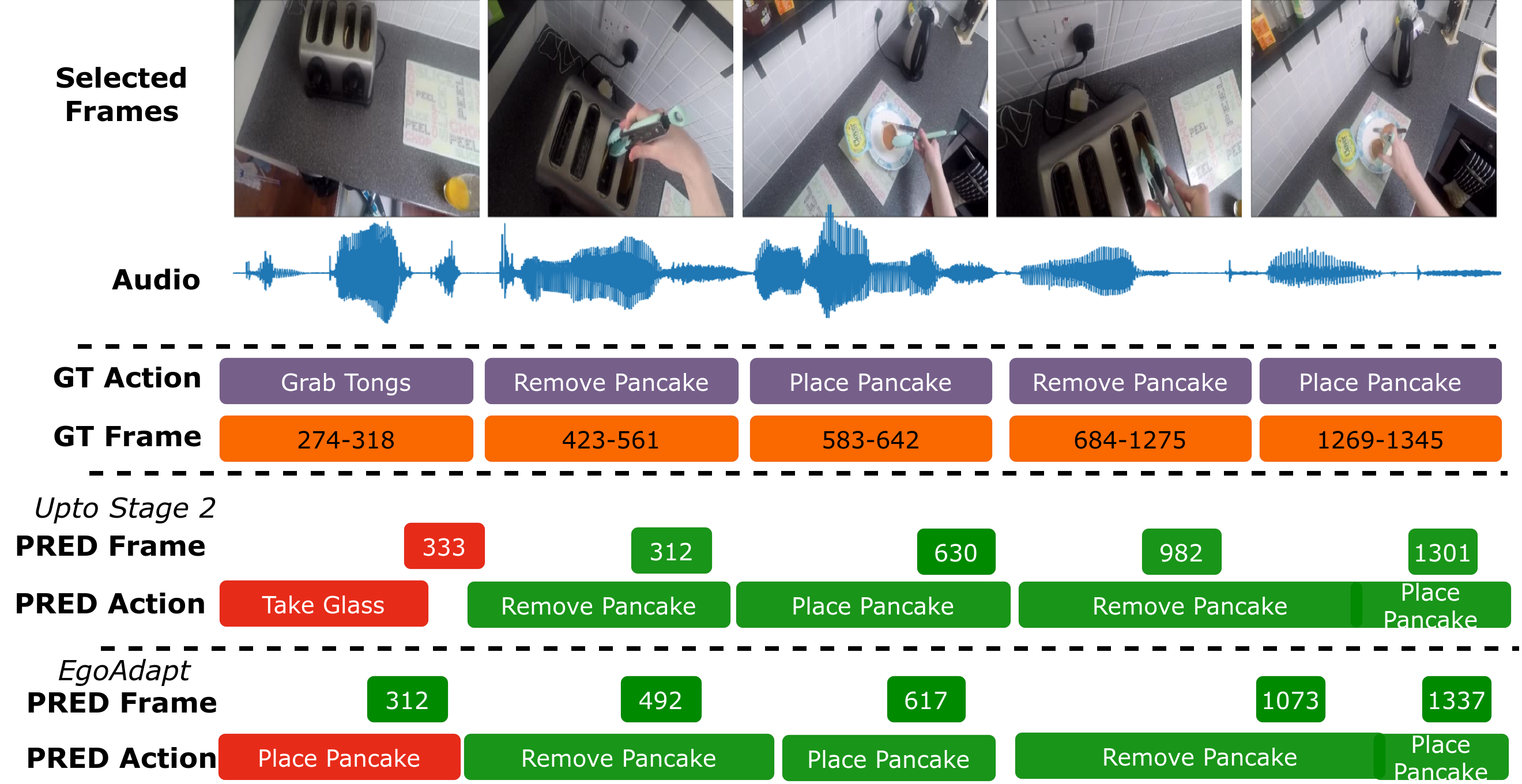}
    \vspace{3mm}
\end{figure*}

\begin{figure*}[h]
    \centering
    \includegraphics[width=\textwidth]{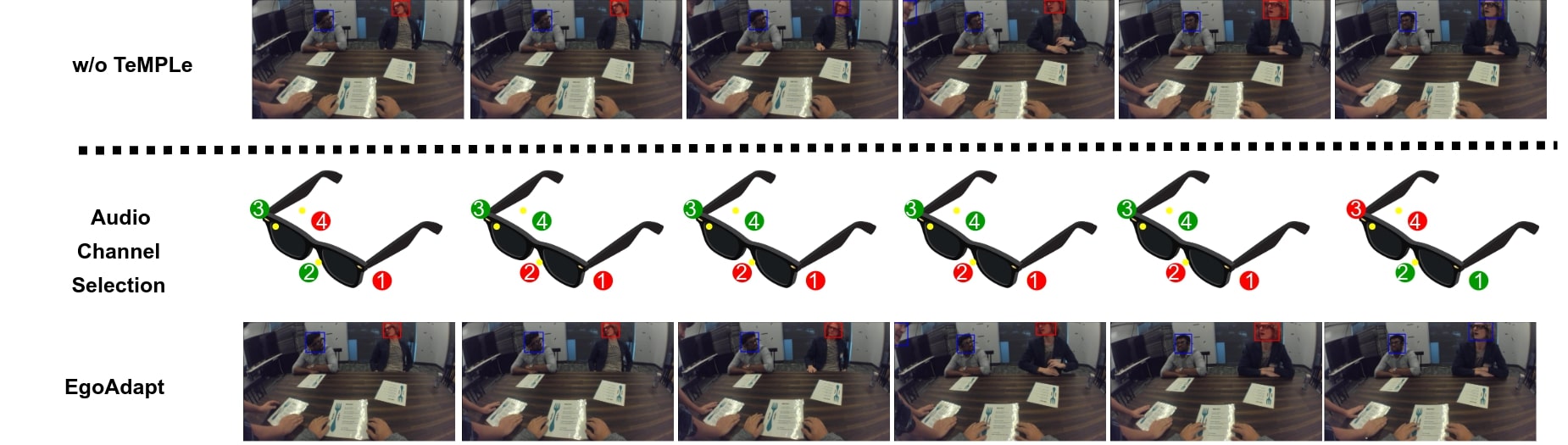}
    \vspace{3mm}
\end{figure*}

\begin{figure*}[h]
    \centering
    \includegraphics[width=\textwidth]{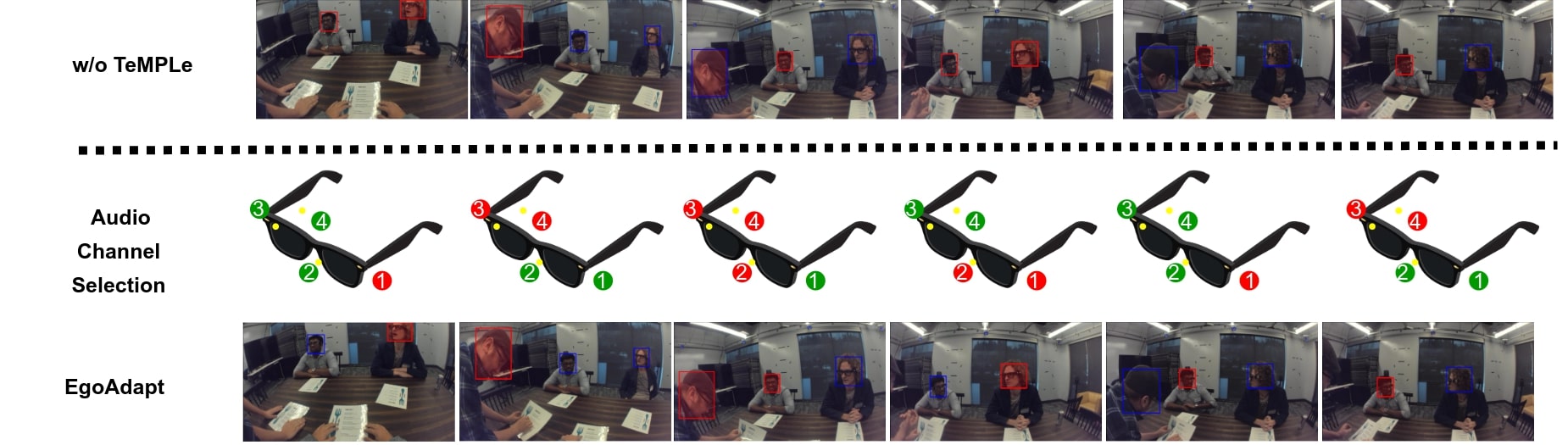}
    \vspace{3mm}
\end{figure*}

\begin{figure*}[h]
    \centering
    \includegraphics[width=\textwidth]{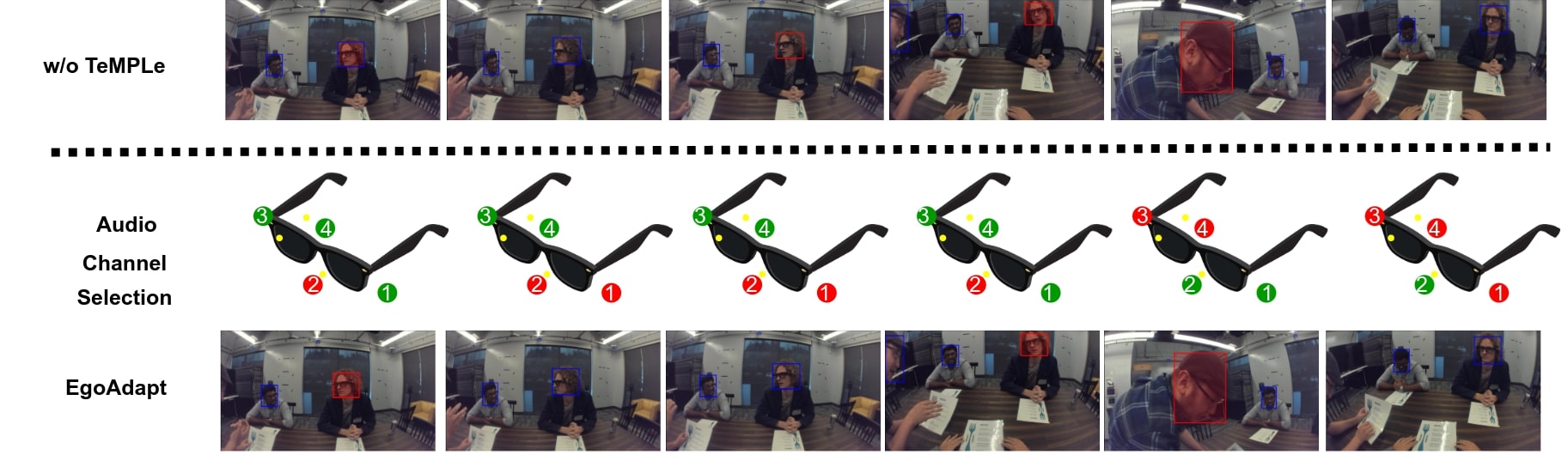}
    \vspace{3mm}
\end{figure*}

\begin{figure*}
    \centering
\includegraphics[width=0.7\textwidth]{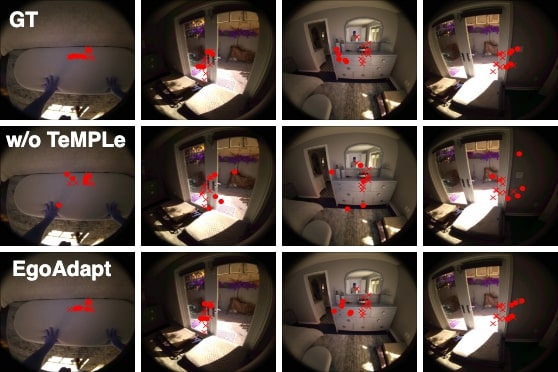}
\end{figure*}

\begin{figure*}[ht!]
    \centering
\includegraphics[width=0.7\textwidth]{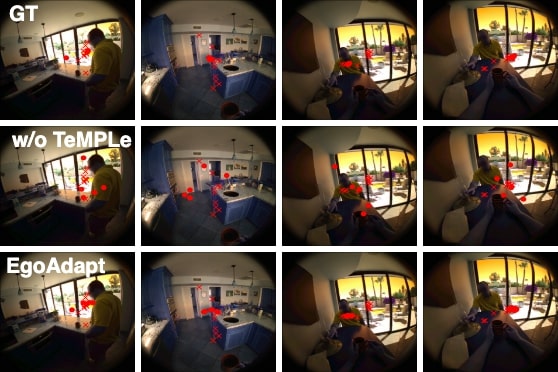}
    \caption{\textbf{Qualitative examples of egocentric behavior anticipation on the AEA Dataset.} Cross/circle symbols denote previous/anticipated behaviors.}
    \label{fig:qual_aea}
    \vspace{-4mm}
\end{figure*}

\clearpage
\raggedbottom
\clearpage

\end{document}